\documentclass[letterpaper]{article} 
\usepackage[]{aaai25}  
\usepackage{times}  
\usepackage{helvet}  
\usepackage{courier}  
\usepackage[hyphens]{url}  
\usepackage{graphicx} 
\usepackage{natbib}  
\usepackage{caption} 
\usepackage{multirow}

\usepackage{algorithm}
\usepackage{algorithmic}
\usepackage{amsmath,amssymb}

\usepackage{booktabs}

\usepackage{newfloat}
\usepackage{listings}
\DeclareCaptionStyle{ruled}{labelfont=normalfont,labelsep=colon,strut=off} 
\floatstyle{ruled}
\newfloat{listing}{tb}{lst}{}
\floatname{listing}{Listing}
\title{Generative Topological Networks}
\author {
    Alona Levy-Jurgenson\textsuperscript{\rm 1,2,3},
    Zohar Yakhini\textsuperscript{\rm 1,2}
}
\affiliations {
    \textsuperscript{\rm 1}Computer Science Department, Reichman University\\
    \textsuperscript{\rm 2}Computer Science Department, Technion -- Israel Institute of Technology\\ 
    \textsuperscript{\rm 3}Department of Statistics, University of Oxford\\
    Correspondence: alona.jurgenson@stats.ox.ac.uk 
}
\usepackage{bibentry}
\usepackage{xcolor}

\begin{document}

\maketitle

\begin{abstract}
Generative methods have recently seen significant improvements by generating in a lower-dimensional latent representation of the data. However, many of the generative methods applied in the latent space remain complex and difficult to train. Further, it is not entirely clear why transitioning to a lower-dimensional latent space can improve generative quality. In this work, we introduce a new and simple generative method grounded in topology theory – \textit{Generative Topological Networks (GTNs)} – which also provides insights into why lower-dimensional latent-space representations might be better-suited for data generation. GTNs are simple to train – they employ a standard supervised learning approach and do not suffer from common generative pitfalls such as mode collapse, posterior collapse or the need to pose constraints on the neural network architecture. We demonstrate the use of GTNs on several datasets, including MNIST, CelebA, CIFAR-10 and the Hands and Palm Images dataset by training GTNs on a lower-dimensional latent representation of the data. We show that GTNs can improve upon VAEs and that they are quick to converge, generating realistic samples in early epochs. Further, we use the topological considerations behind the development of GTNs to offer insights into why generative models may benefit from operating on a lower-dimensional latent space, highlighting the important link between the intrinsic dimension of the data and the dimension in which the data is generated. Particularly, we demonstrate that generating in high dimensional ambient spaces may be a contributing factor to out-of-distribution samples generated by diffusion models. We also highlight other topological properties that are important to consider when using and designing generative models. Our code is available at: https://github.com/alonalj/GTN
\end{abstract}

%

\section{Introduction}

\begin{figure*}[h]
\centering
\includegraphics[width=1\textwidth]{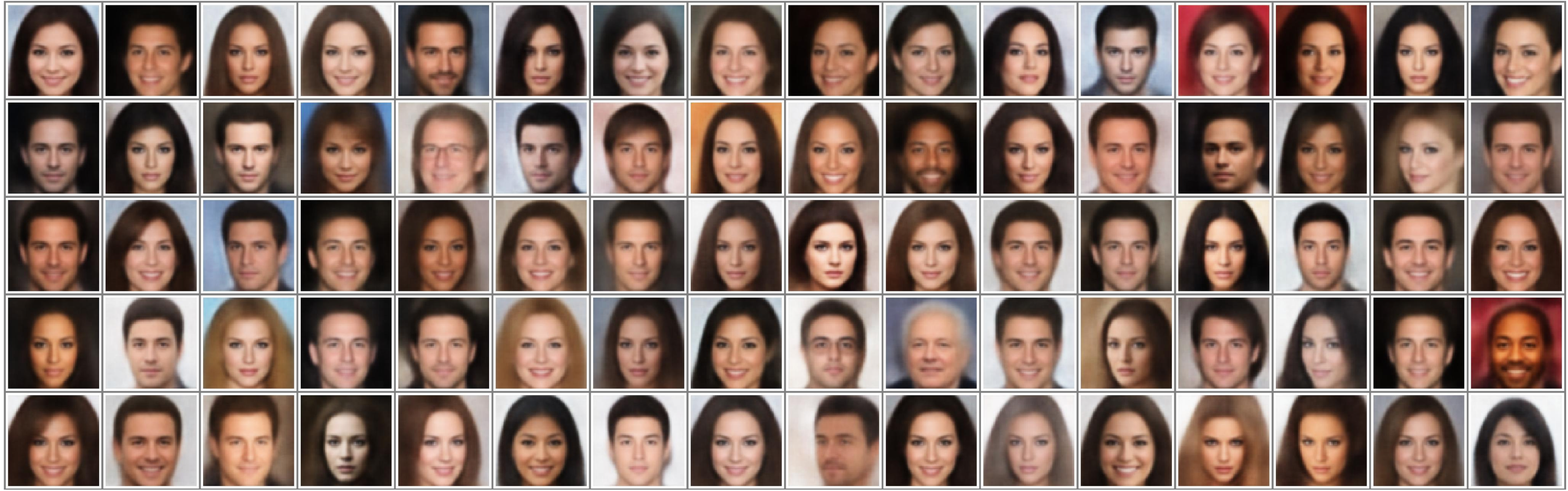}
\caption{Samples generated 
by a GTN trained on a latent representation of CelebA $64\times 64$ with latent dimension $d=100$.}
\label{fig:celeba_grid}
\end{figure*}

Deep generative models such as Generative Adversarial Networks (GANs) \cite{goodfellow2020generative}, Variational Autoencoders (VAEs) \cite{kingma2013auto}, Energy-Based Models (EBMs) \cite{lecun2006tutorial, ngiam2011learning}, 
normalizing flows (NFs) \cite{rezende2015variational}, and diffusion models \cite{sohl2015deep}, have demonstrated remarkable capabilities for generating samples based on training data distributions \cite{kang2023scaling, ho2020denoising, ramesh2021zero, ho2019flow++, kingma2018glow, kingma2016improved, reed2017parallel, van2017neural, ramesh2021zero, onken2021ot}.

Many image-generation methods like diffusion are often applied in pixel-space \cite{ho2020denoising, sohl2015deepWithAppx}. More recently, generative methods have seen great improvements in generative quality by transitioning to a lower-dimensional latent representation of the data \cite{rombach2022high}. Although training in a lower-dimensional latent space reduces the computational burden, the methods used to generate in the latent space remain complex and computationally expensive.
Further, besides the obvious training efficiency, it is not entirely clear why transitioning to a lower-dimensional latent representation of the data leads to improved generative quality compared to pixel-space, or by how much the dimension should be reduced. 

In this work we introduce a new class of generative models -- Generative Topological Networks (GTNs) -- that provides a simple    approach for generating in the latent space and that sheds light on why transitioning to a lower-dimensional latent space improves generative quality.

We begin by introducing GTNs. Given a training set of samples (e.g. images) and a tractable source distribution (e.g. Gaussian), GTNs learn to approximate a continuous and invertible function $h$ such that, given a $y$ sampled from the source distribution, $h(y)$ is a sample representing the training data distribution.
GTNs are reminiscent of NFs, which aim to transform one distribution into another using a sequence of invertible and differentiable maps. 
NFs, however, pose specific constraints on the network architecture. In contrast, GTNs do not pose any constraints, and are fully operational using a simple vanilla feedforward architecture.

From a \textbf{practical} perspective, 
GTNs are simple to train, requiring only a single, vanilla, feedforward architecture trained using standard supervised learning. This allows GTNs to avoid issues like mode collapse or posterior collapse faced by GANs and VAEs;
to circumvent the intricacies of training more complex architectures such as those employed by diffusion and GANs;
and to avoid posing constraints on the structure of the neural network, as in NFs.
These advantages manifest in our experiments -- with realistic samples obtained at early epochs using a single T4 GPU.

From a \textbf{theoretical} perspective, GTNs provide guarantees and properties that are desirable in the context of generative models. These include: learnability (via the universal approximation theorem), continuity (for continuous interpolations -- Figure \ref{fig:interp}), bijectivity (for diversity and coverage of the data distribution --  Figure \ref{fig:intro_methods}) and  properties that serve as guiding topological principles that can inform the design of generative methods (see the Method section and the swiss-roll example). As an example, we use some of the topological principles behind GTN to demonstrate that generating with diffusion without transitioning to a lower-dimensional latent space can lead to many out-of-distribution samples. We also discuss why this is important for other methods too.


The remainder of the paper is structured as follows:  In the Method section we first develop the theory behind GTNs for the $1$-dimensional (1D) case and use it to accurately generate samples from an intrinsically 1D swiss-roll represented in a 2D ambient space. Here, we  emphasize the significance of the intrinsic dimension of the data for generative models and demonstrate that diffusion struggles to accurately generate the 1D swiss-roll in the original 2D ambient space. We then extend the theory behind GTNs from the 1D case to higher dimensions and use it to accurately generate samples from the multivariate uniform distribution. Finally, we apply GTNs to real datasets represented in a lower-dimensional latent space obtained by autoencoders. We show that GTNs generate samples resembling reconstructed data and that they improve upon VAEs both quantitatively and qualitatively. We conclude with Related Work and Discussion sections where we elaborate on how GTNs compare to other methods from a practical perspective, and demonstrate extensions of GTNs to more complicated data.

\section{Method} \label{sec:method}
We first develop the method in the simple case of a $1$D space, and proceed to generalize to higher dimensions.

\subsection{1-Dimension}\label{sec:1dim}
\begin{figure}
\centering
    \includegraphics[width=0.8\columnwidth]
    {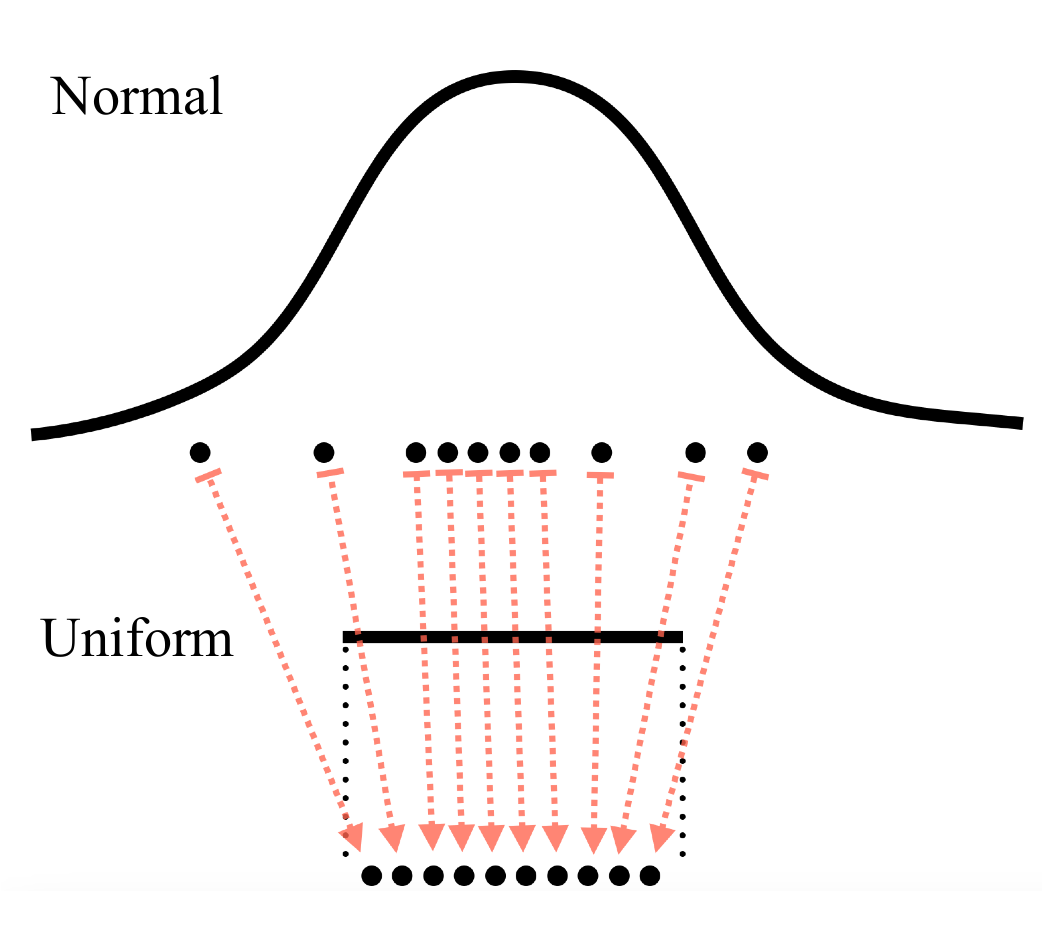}
\caption{Illustration of the mapping produced by $h$ and of the labeling process for training its approximation $\hat{h}$, where $Y$ is normally distributed and $X$ is uniformly distributed. A point $y$ from the normal sample is labeled with the unique point $x_y$ from the uniform sample that has the same empirical CDF value as $y$. }
\label{fig:labeling}
\end{figure}

Consider a continuous random variable $X$ with values in $\mathbb{R}$.
We wish to generate samples from $X$ without knowing its distribution. One solution would be to sample from a known and tractable distribution, such as the standard normal distribution, and then to apply a function that maps this sample to a corresponding sample from $X$. Diffusion models attempt to approximate such a mapping through gradual stochastic manipulations of the standard normal sample back to a sample from $X$. We will show that, under certain general conditions, such a mapping can be explicitly defined, providing a simple deterministic function which we will denote as $h$ (and which we illustrate in Figure \ref{fig:labeling}). We will show that $h$ is in fact a homeomorphism -- it is continuous, invertible and has a continuous inverse (see Definition \ref{def_homeo}). This has significant implications, as we will soon explain.


\subsubsection{Defining $h$.}\label{sec:definition_h}
In this section we will define the aforementioned function $h$ that transforms one distribution into another.
We will also prove that $h$, under fairly general conditions, possesses certain properties that are desirable in the context of generative models by proving that it is a homeomorphism (Theorem \ref{thm}). For example, we will see that $h$ is  bijective, so that each sample $y$ is mapped precisely to one sample $x_y=h(y)$, generating different samples for different $y$, and guaranteeing that each sample $x$ has a sample $y$ that generates it. We will also discuss other useful consequences of $h$ being a homeomorphism. 
We begin by defining the term \textit{homeomorphism} in our context and proceed to defining $h$ in Theorem \ref{thm}.

\textbf{Definition 2.1}\label{def_homeo} \textit{(Homeomorphism for $\mathbb{R}^n$)}. 
Let $S, T$ be two subsets of $\mathbb{R}^n$. A function $h : S \to T$ is a \textbf{homeomorphism} if: (1) $h$ is continuous; (2) $h$ is bijective; (3) $h^{-1}$ is continuous.
When such an $h$ exists, then $S$ and $T$ are called \textbf{homeomorphic}. 


\textbf{Theorem 2.1}\label{thm}
Let $X$ and $Y$ be random variables 
that have continuous probability density functions (pdfs) $f_X$, $f_Y$ and supports $S_X$, $S_Y$ that are open intervals in $\mathbb{R}$. 
Denote the corresponding cumulative distribution functions (CDFs) as $F_X$ and $F_Y$. Define:
\begin{equation} \label{h_appx_open_interval}
\begin{split}
h : S_Y & \to S_X \\
      h(y)  &=  {F_X}|_{S_X}^{-1}({F_Y}|_{S_Y}(y))
\end{split}
\end{equation}
Then: (1) $h$ is well-defined; 
(2) $h$ is a homeomorphism.

Note that the requirement that $S_X, S_Y$ are open intervals can be adapted to a union of open intervals -- see Appendix \ref{appx_propertyTopo}. The proof of Theorem \ref{thm} is in Appendix \ref{appx_thm}. 

The simple special case of $f_X, f_Y >0$ in $\mathbb{R}$ illustrates the key ideas of Theorem \ref{thm}. In this case we have that: 

\begin{equation} \label{eq1}
\begin{split}
h : \mathbb{R} & \to \mathbb{R} \\
   h(y)  &=  F_X^{-1}(F_Y(y))
\end{split}
\end{equation}




For simplicity of exposition, we continue with this special case 
but note that adapting the results to the general case is a matter of technicality. Namely -- $f_X, f_Y >0$ on $S_X, S_Y$ respectively. Combined with $S_X$ and $S_Y$ being open intervals, we conclude that the restricted CDFs ${F_X}|_{S_X}, {F_Y}|_{S_Y}$ are continuous and strictly monotonically increasing on $S_X, S_Y$, which is the main observation needed in order to generalize. 




On the many advantages of $h$ being a homeomorphism in the context of generative models see Appendix \ref{sec:significance}.

\begin{figure*}[h]
\centering
\includegraphics[width=\textwidth]{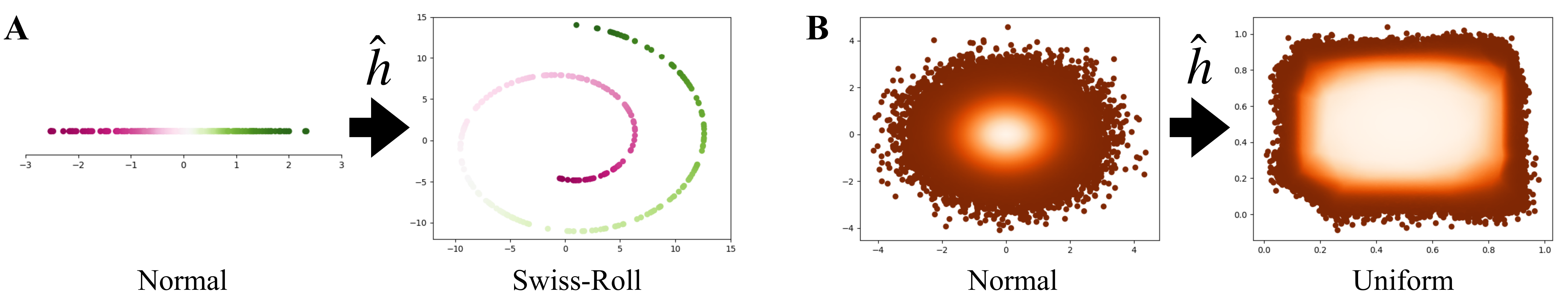}
\caption{\textbf{(A)} Test results for a GTN $\hat{h}$ trained to map from $Y\sim \mathcal{N}(0,1)$ to the swiss-roll parameter. The color indicates which point in the normal sample was mapped to which point in the swiss-roll. \textbf{(B)} Test results for a GTN $\hat{h}$ trained to map from $Y\sim \mathcal{N}(\textbf{0},\textbf{I})$ to $X\sim U\big((0,1)\times(0,1)\big)$. The color is based on the normal sample (left): for each $y$ in the normal sample, $\hat{h}(y)$ has the same color as $y$ so that the figure on the right shows how the normal sample was stretched to a uniform distribution.  }
\label{fig:intro_methods}
\end{figure*}

\subsubsection{Learning $h$ Using a Neural Network $\hat{h}$ and Generating Samples with $\hat{h}$.} 
As explained in the previous subsection, one consequence of $h$ being a homeomorphism is that it can be approximated by a feedforward neural network $\hat{h}$. 
If we had known both $F^{-1}_X$ and $F_Y$ 
then we would be able to easily generate labels for each $y\in Y$: denoting the label of a given $y \in Y$ as $x_y$, we would set $x_y = h(y)$, meaning:

\begin{equation} \label{Eq:x_y}
x_y =h(y)= F_X^{-1}(F_Y(y))  
\end{equation}



Although we do not know $F_X$, $F_Y$, we do have access to samples from $X$ and $Y$ which we can use to approximate $x_y$.
From Eq. \ref{Eq:x_y} we see that $x_y$ satisfies $F_X(x_y)=F_Y(y)$, meaning $x_y$ is the unique $x \in X$ that matches the percentile of $y$.
We can use this observation to approximate $x_y$ by replacing $F_X$ and $F_Y$ with the empirical CDFs as follows.

Let $D_X := \{x_1, \dots, x_n\}$ and $D_Y := \{y_1, \dots y_n\}$ be $n$ observed values of $X$ and $Y$, respectively. Labels are obtained easily by sorting $D_X$ and $D_Y$ to obtain $D_X^\text{sorted}$ and $D_Y^\text{sorted}$ and labeling each $y\in D_Y^\text{sorted}$ with the corresponding $x \in D_X^\text{sorted}$ (i.e. assuming $D_X, D_Y$ are sorted, $y_i$ is assigned $x_i$). This is illustrated in Figure \ref{fig:labeling}.

The loss function is the MSE: $$\frac{1}{n}\sum_{i=1}^n||\hat{h}(y_i)-x_{y_i}||^2$$

Generating samples from $X$ is now straightforward -- we sample $y\in Y$ and compute $\hat{h}(y)$.

\subsubsection{Example: Swiss-Roll.}\label{sec:swiss-roll}
\begin{figure}[ht!]
\centering
\includegraphics[width=0.5\columnwidth]{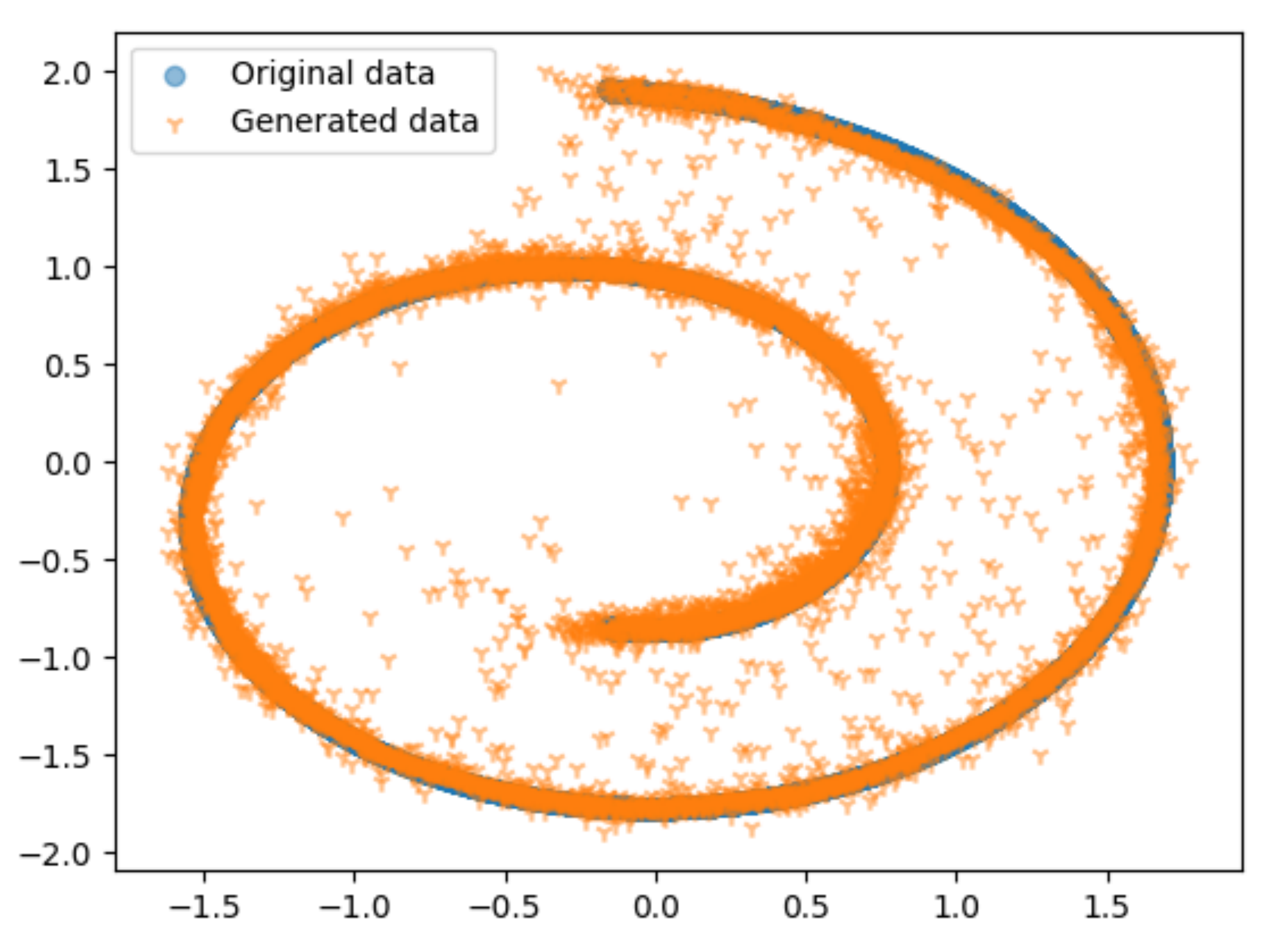}
\caption{Samples generated by a diffusion model when trained using a 2D Gaussian and a 2D representation of the 1D swiss-roll. Many of the generated samples fall out-of-distribution. We use the diffusion model provided by \cite{toyDiffusion}, adapting only the input. The plot format is also by \cite{toyDiffusion}.}
\label{fig:diffusion_cant_do_1d_swiss}
\end{figure}

In this example, we show that samples from the swiss-roll can be easily generated from a $1$D normal distribution with visibly near-perfect accuracy (Figure \ref{fig:intro_methods} (A)) using a GTN. 
This would not be as easy using a $2$D normal distribution, 
as explained next, and as shown by an attempt to do so with diffusion in Figure \ref{fig:diffusion_cant_do_1d_swiss}. 

The swiss-roll is a $1$D manifold in a $2$D space $\mathbb{R}^2$ (generated using a single, 1-dimensional, continuous random variable $\theta$). 
Being a $1$D manifold, it cannot be homermorphic to the support of a $2$D normal distribution ($\mathbb{R}^2$) since homeomorphic manifolds must have the same dimension. Therefore, there would be no hope of learning a homeomorphism that maps a $2$D Gaussian random variable to the swiss-roll.   
However, the swiss roll \textit{is} homeomorphic to $\mathbb{R}$ -- the support of a $1$D standard normal distribution. In fact, since $\theta$ is a random variable that satisfies the conditions in Theorem \ref{thm}, it is homeomorphic to $\mathbb{R}$ via $h$, which we can learn using $\hat{h}$. 
The fact that there is no hope of learning a homeomorphism in a higher-dimensional space than the intrinsic dimension is also important for other generative methods like NFs, which use a diffeomorphism (a type of homeomorphism) as a generator, and it may also help explain why latent diffusion models have shown improvements over pixel-space diffusion models \cite{rombach2022high}.

%

To train $\hat{h}$, we created a dataset of $n=50,000$ samples from $\theta$ (Appendix \ref{appx_swiss_1d_method}), denoted $D_X$. We sampled $n$ samples from a $1$D standard normal distribution to obtain $D_Y$. We labeled each $y_i \in D_Y$ with its $x_{y_i} \in D_X$ as defined in the previous section. We used a standard feedforward neural network as $\hat{h}$ ($4$ layers of width $6$; details in Appendix Table \ref{tab:architectures_h_hat}). 
Figure \ref{fig:intro_methods} (A) shows the result of testing the trained model $\hat{h}$ on a set of new samples $y_1,\dots ,y_k$ drawn from $\mathcal{N}(0,1)$ (each point is obtained by predicting $\hat{\theta}_i := \hat{h}(y_i)$ and applying the formula for the swiss roll to $\hat{\theta}_i$).  

For comparison, 
we refer the reader to Figure 1 (middle) in \cite{sohl2015deep} where a trained diffusion model was applied to points sampled from a standard $2$D normal distribution. 
The points are gradually and stochastically moved back towards the swiss-roll manifold, but the result contains points that fall outside of this manifold. With $\hat{h}$, the  generated points lie entirely within the data manifold. In fact, this holds true early in the training process, as shown in Appendix Figure \ref{fig:swiss_epochs}.  

Besides demonstrating that $\hat{h}$ can serve as an accurate generative model, this example also emphasizes an important point -- namely, that if we want to learn such a homeomorphism $h$, we might need to reduce the dimensionality of the data if it is not already in its intrinsic dimension (for the swiss-roll this was $1$). This observation will guide us in later sections.


\subsection{Higher Dimensions}\label{sec:higher_dim}
To generalize to more than one dimension, we reduce to the $1$D case by considering lines that pass through the origin. Briefly, we take the random variables obtained by restricting $X$ and $Y$ to each line, and apply the homeomorphism defined for the $1$D case to these random variables. 
We begin with a formal setup that is very similar to the $1$D case. 

\subsubsection{Defining $h$.}

Let $X=(X_1,\dots,X_d), Y=(Y_1,\dots ,Y_d)$ be multivariate random-variables with continuous joint probability density functions $f_X$, $f_Y$, and with supports $S_X$, $S_Y$, each of which is a product of $d$ open intervals in $\mathbb{R}$ (see Appendix \ref{appx_propertyTopo} to accommodate for other supports). For example, $X$ could be uniformly distributed with support $S_X = (0,1)\times(0,1)$ and $Y$ could be normally distributed with support $S_Y = \mathbb{R}\times\mathbb{R}$.


Consider first rotation invariant distributions (e.g., the standard multivariate normal). That is, assume that for every rotation matrix $R$ and every $x \in \mathbb{R}^d$, $f_X(R x) = f_X(x)$, and similarly for $f_Y$.  For simplicity, also assume that $S_Y, S_X = \mathbb{R}^d$. We can now define the following homeomorphism:

\begin{equation} \label{h_multivar}
\begin{split}
h &: \mathbb{R}^d \to \mathbb{R}^d \\
   h&(y)  = \begin{cases} 
          h_1(||y||)\frac{y}{||y||}, \quad y\neq 0\\
          0, \quad y = 0 \\
       \end{cases}  
\end{split}
\end{equation}

    where $h_1:(0,\infty) \to (0,\infty)$ is the $1$D homeomorphism applied to the random variables $||Y||, ||X||$.

    We prove that this is indeed a homeomorphism in Appendix \ref{appx_thm2}. Note that this generalizes the $1$D case since for $d=1$ we get $y \mapsto h_1(y)$.
    Intuitively, $h$ can be seen as shrinking or stretching $y$ to the unit vector in $y$'s direction ($y/||y||)$, and then shrinking or stretching it to reach the $x_y := h(y)$ that has the same ranked distance as $y$ on the line (by multiplying it by $h_1(||y||)$). 
    More precisely, $h_1$ produces $x_y$'s distance from the origin so that it has the same quantile as $y$'s distance from the origin (when measured with respect to the random variables obtained by restricting $Y, X$ to the line segment from the origin in $y$'s direction). 
    Another way of thinking about this is provided in Appendix \ref{appx_another_way_lines}.  

    For more complicated distributions, it may be difficult to explicitly define a similar $h$ since $h_1$ would depend on the line. Nevertheless, the simpler case above can guide us on how to train a neural network to learn one (as accomplished in the non-rotation-invariant 2D uniform case in Figure \ref{fig:intro_methods} (B)).

    


\subsubsection{Learning $h$.}
\begin{figure}[ht!]
\centering
\includegraphics[width=0.8\columnwidth]{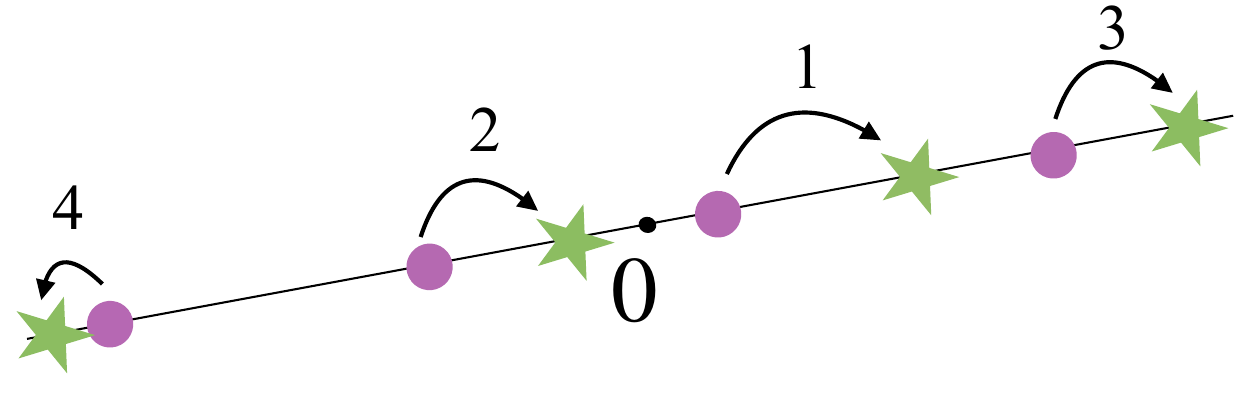}
\caption{Illustration of Algorithm \ref{algorithm_labeling}. Numbers reflect the order of matching $y$ (circles) with $x_y$ (stars).}
\label{fig:illustration_algorithm_four_points}
\end{figure}

Let $D_X$ and $D_Y$ be observed values from $X$ and $Y$. 
Ideally, we would use the $1$D labeling scheme on each line that passes through the origin by using $h_1$ on both sides of the line. In practice, however, the probability that any given line contains points from either $D_X$ or $D_Y$ is negligible.
We therefore approximate this labeling approach using cosine-similarity. Specifically, we 
match $y$ with the $x$ that is closest in terms of both cosine-similarity (so that both $y, x_y$ are as close as possible to being on the same line and in the same direction) and distance from the origin (to approximate $h_1(||y||)$).
This is formally described in Algorithm \ref{algorithm_labeling} and illustrated in  Figure \ref{fig:illustration_algorithm_four_points}. 
To see why Algorithm \ref{algorithm_labeling} approximates $h$, 
consider a line with an infinite sample from $X$ and $Y$ in both directions from the origin (see Figure \ref{fig:illustration_algorithm_four_points}). The maximal cosine-similarity over all $x\in D_X$ and a $y$ on this line is $1$, so the algorithm returns an $x_y$ that is on this line and in $y$'s direction. Because $D_X, D_Y$ are sorted by distance from the origin, ties in cosine-similarity are broken by distance from the origin, so that the first $y$ is matched with the first $x$, the second $y$ with the second $x$ etc. as in the $1$D case, i.e. $||x_y||$ approximates $h_1(||y||)$.
Appendix \ref{appx_rationaleAlgo} gives further intuition. 





Using the labeled samples from Algorithm \ref{algorithm_labeling}, we train $\hat{h}$ -- a  feedforward neural network -- using MSE as the loss function. We then use $\hat{h}$ to generate new $X$ instances just as in the $1$D case, formally described in Algorithm \ref{algorithm_sampling}.

\begin{algorithm}[tb]
\caption{Labeling}
\label{algorithm_labeling}
\textbf{Input}: \\$D_X=\{x_1, \dots, x_n\}$ \\ $ \quad\quad\quad D_Y=\{y_1, \dots, y_n\} \text{ sampled from } \mathcal{N}(\textbf{0},\textbf{I})$ \\
\textbf{Output}: \textit{res} 
\begin{algorithmic}[1] 
\STATE $\textit{res} \leftarrow []$
\STATE $D_X^{\text{sorted}}, D_Y^{\text{sorted}} \leftarrow \text{ sort } D_X, D_Y \text{ ascending by $||\cdot||_2$}$
    \STATE \textbf{while}  $D_Y^{\text{sorted}} \neq \emptyset$:
    \STATE $\quad\quad y \leftarrow D_Y^{\text{sorted}}[0]$
    \STATE $\quad\quad x_y \leftarrow \operatorname*{arg\,max}_{x\in D_X^{\text{sorted}}} \textit{cosine\_sim}(x,y)$
    \STATE $\quad\quad D_Y^{\text{sorted}} \leftarrow D_Y^{\text{sorted}}[1\dots n]$
    \STATE $\quad\quad D_X^{\text{sorted}} \leftarrow D_X^{\text{sorted}}\setminus \{x_y\}$
    \STATE $\quad\quad\textit{res.append}\big((y, x_y)\big)$
    \STATE $\textbf{return}\textit{ res}$
\end{algorithmic}
\end{algorithm}

\begin{algorithm}[tb]
\caption{Sampling}
\label{algorithm_sampling}
\textbf{Input}: $\hat{h}$ trained on \textit{res} (see Alg. 1)
\begin{algorithmic}[1] 
\STATE $y \leftarrow \text{ sample from } \mathcal{N}(\textbf{0},\textbf{I})$
\STATE \textbf{return} $\hat{h}(y)$
\end{algorithmic}
\end{algorithm}

\subsubsection{Example: 2-Dimensional Uniform Distribution.}\label{sec:uniform2d}
Figure \ref{fig:intro_methods} (B) shows a sample generated using Algorithm \ref{algorithm_sampling} after applying the method to the multivariate uniform distribution $X \sim U(0,1)\times U(0,1)$. 
Specifically, we created $D_X$ by sampling $n=100,000$ points from $X$, and created $D_Y$ by sampling the same number of points from $\mathcal{N}(\textbf{0},\textbf{I})$. We then applied Algorithm \ref{algorithm_labeling} to $D_X, D_Y$ to generate labeled data. We trained a standard feedforward neural network $\hat{h}$ (6 layers, width 6; details in Appendix Table \ref{tab:architectures_h_hat}) and used it to generate new samples according to Algorithm \ref{algorithm_sampling}. The model was trained until convergence on a separately generated validation set with $n=10,000$. The colors in both images in Figure \ref{fig:intro_methods} (B) are based on the distance of the points from the origin in the Gaussian sample (the left image in (B)), so that the image on the right reflects where each point in the Gaussian sample was predicted to 'move' to by $\hat{h}$. 

\begin{figure*}[h]
\centering
\includegraphics[width=\textwidth]{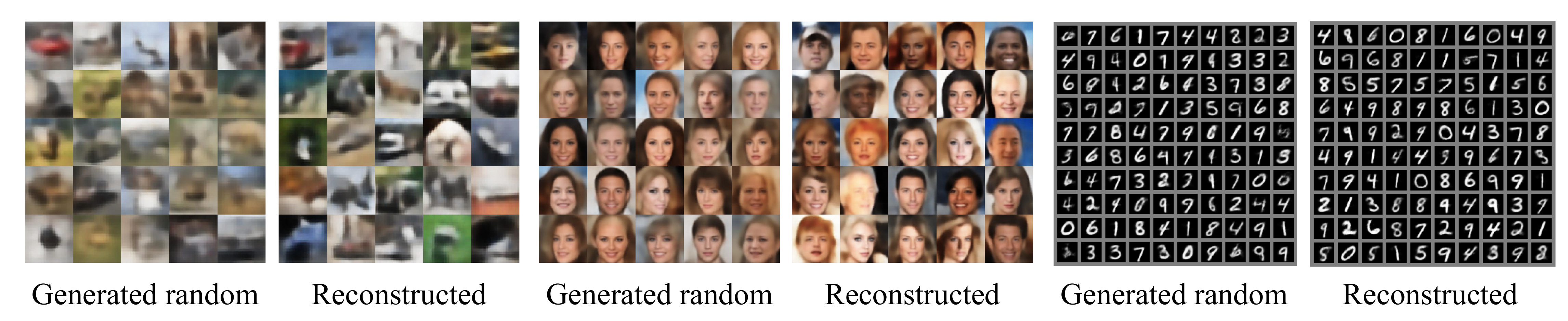}
\caption{Random generated and reconstructed images for CIFAR-10 (left), CelebA (centre) and MNIST (right). Each of the generated samples is the decoded $\hat{h}(r)$ for a random $r \sim \mathcal{N}(\textbf{0},\textbf{I})$) and each of the reconstructions is the decoded vector of a random real image. }
\label{fig:cifar_celeb_mnist}
\end{figure*}

\section{Experiments} \label{sec:experiments}

In the previous section, we demonstrated how to learn $\hat{h}$ on synthetic data -- the swiss-roll and the multivariate uniform distribution. In this section, we will demonstrate how to apply $\hat{h}$ to images. 

\subsubsection{Setup.} In light of the discussion in Section \ref{sec:swiss-roll} on using the intrinsic dimension (ID) of the data, we use autoencoders to represent each dataset in a lower dimensional latent space before training the GTN $\hat{h}$. 
Each dataset was trained separately (with its own autoencoder and $\hat{h}$). 
Unless mentioned otherwise, all experiments used the same vanilla autoencoder architecture adapted to different latent dimensions (the encoder consisted of two convolution layers with ReLU activation followed by a fully connected layer).
For $\hat{h}$ we use a standard feedforward neural network with the width and depth depending on the data. Architecture and training details can be found in Appendix \ref{appx_architectures_and_training} and Table \ref{tab:architectures_h_hat} with further details in our code.


To train $\hat{h}$, we set $X$ to be the latent vectors of the training set (the encoded images), and set $Y$ to be the standard multivariate normal distribution of the same dimension as $X$ (e.g., for a latent dimension of $d=100$ in the autoencoder, $Y$ has dimension $100$). Images were generated using Algorithm \ref{algorithm_sampling} by computing: $autoencoder.decoder\big(\hat{h}(r)\big)$ where $r \sim \mathcal{N}(\textbf{0},\textbf{I})$ (generation occurs in the latent space). 

\subsubsection{Evaluation. } 
We applied our method to MNIST, CIFAR-10, CelebA $64 \times 64$ and the Hands and Palm (HaP) datasets.

We first applied our method to MNIST \cite{lecun1998gradient} using a latent dimension of $d=5$ to allow for comparison with VAE for the same latent dimension (see Figure 5(b) in \cite{kingma2013auto}). Figure \ref{fig:cifar_celeb_mnist} (right) shows random sets of generated images for MNIST, as well as a sample of random reconstructed images for comparison. 

Next we applied the method to CelebA \cite{liu2015deep} (Figure \ref{fig:celeba_grid} and Appendix Figure \ref{fig:celeba_epochs}).
Images were center cropped to $148\times 148$ and resized to $64\times 64$.
We tested latent dimensions of $d \in \{10, 25, 50 ,100\}$, which are within the range of the typical ID estimated for image datasets ($10$-$50$) or close ($100$) \cite{pope2021intrinsic}. We used Inception Score (IS) \cite{salimans2016improved},  a common evaluation metric, as the stopping criteria: after every epoch, the GTN generated $200$ random images for which the IS was evaluated. We used IS instead of validation-set results since the IS continued to improve well beyond the point of plateau on the validation set. Training stopped after $300$ epochs of no improvement in IS. 
For evaluation, we used both IS and Fréchet Inception Distance (FID) \cite{heusel2017gans}. For IS, in addition to the best IS  obtained by the model, we produced the IS for a single random set of $200$ reconstructed images ("recon-IS") since this reflects the best possible IS that we can reasonably hope to achieve. 
Plotting IS and recon-IS by epoch, (Appendix Figure \ref{fig:epoch_to_IS_grid_CelebA}) shows that IS increases throughout the training process, and that recon-IS is either achieved ($d \in \{10, 25\}$) or nearly achieved ($d \in \{50, 100\}$) by the GTN.
For FID, the lowest FID was for $d=100$ with 66.05 and the highest was for $d=10$ with 119.46. FID and IS across all dimensions are reported in Appendix Table \ref{tab:IS_FID}. 

Observing the progress in image generation during training (Appendix Figure \ref{fig:celeba_epochs}), shows that realistic images were already obtained at half the training time (see epoch $276$).
One epoch took just under $1$ minute ($50.6$ seconds on average) on a single T4 GPU, reflecting $9$ hours until the last improvement in IS at epoch $640$ and $<4$ hours to reach epoch $276$. 
The relatively fast convergence raises the question whether the latent space was approximately normal to begin with. Appendix Figure \ref{fig:random_normal_vs_gen}, which compares images generated by decoding random normal vectors with images generated using the GTN, demonstrates that this is not the case.

\begin{figure}[h]
\centering
\includegraphics[width=1\columnwidth]{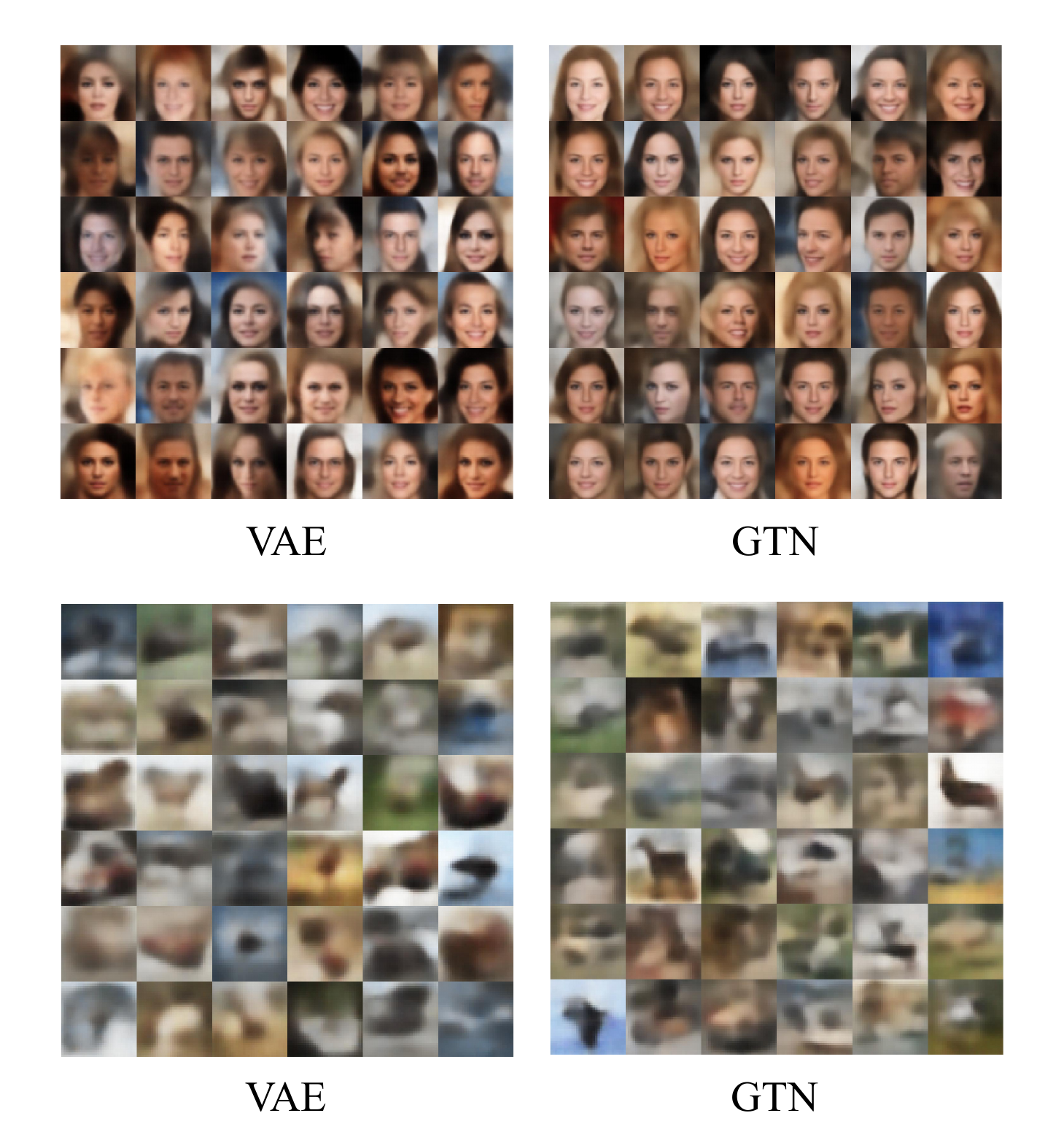}
\caption{Controlled comparison of GTN with VAE \cite{kingma2016improved} on CelebA and CIFAR-10 -- randomly generated images from both.}
\label{fig:controlled_comp_celeba_cifar_long}
\end{figure}

Next, we designed a controlled experiment to compare our method to the closely-related VAE on both CelebA and CIFAR-10. We used the vanilla VAE suggested for CelebA from \cite{vanillaVAE} adapting only the latent dimension, and trained two autoencoders: (1) The suggested VAE and; (2) A vanilla autoencoder later used to train a GTN on the learned latent representations. The transition from VAE to vanilla autoencoder included only the necessary adaptations, so that all remaining architectural considerations and hyper-parameters were identical between the two autoencoders. We then used the latent representations from the vanilla autoencoder to train a GTN, where the GTN had the same architecture as in the earlier experiments. No tuning of any kind was performed, including of the random seed (which was set once at the very start of the entire project). Particularly, the GTN architecture and training settings (for both CelebA and CIFAR-10) were unchanged from the previous CelebA experiments (except for the dimensions of the input, output and latent space). Figure \ref{fig:controlled_comp_celeba_cifar_long} shows a side-by-side comparison of randomly generated images from both methods. Table \ref{tab:comp_celeba_cifar_controlled} provides the FID and IS results, including for reconstruction quality (i.e. recon-FID and recon-IS, which compare reconstructions to real images). A comparison with reconstructions is also shown in Figure \ref{fig:cifar_celeb_mnist}. Although GTN outperforms VAE on both CelebA and CIFAR-10, the CIFAR-10 results are overall worse compared to the CelebA results. This is likely since: (a) the autoencoder settings were originally suggested for CelebA; and (b) no fine-tuning or architecture search was performed for the GTN when transitioning from the CelebA experiment to CIFAR-10. For additional context with other methods, we also provide Appendix Tables \ref{tab:comp_celeba_many} and \ref{tab:comp_cifar_many}.

\begin{table*}
    \centering
    \begin{tabular}{ ccccccc }
    \toprule
    \textbf{Data} & \textbf{Method} & \textbf{IS $\uparrow$} & \textbf{recon-IS $\uparrow$} &  \textbf{FID $\downarrow$}  & \textbf{recon-FID $\downarrow$}\\
     \midrule
          \multirow{2}*{CelebA} & VAE  & 1.0 & 1.0 & 94.66 & 83.48 \\ 
     & GTN & 1.90 & 1.94 & 73.18 & 67.08  \\ \midrule
       \multirow{2}*{CIFAR-10} & VAE & 1.77 & 1.90 & 286.49 & 277.48 \\ 
& GTN & 2.07 & 2.24 & 238.62 & 181.53 \\
     \bottomrule
    \end{tabular}
    \caption{Controlled comparison of GTN with VAE \cite{kingma2016improved} on CelebA and CIFAR-10. 
    The VAE result was obtained by training the vanilla VAE architecture supplied in \cite{vanillaVAE}. Prior to training the GTN we trained a new autoencoder that employs the same architectures as the vanilla VAE, with only the necessary adaptations to transition from VAE to a vanilla autoencoder. GTN is trained on the latent representations provided by this autoencoder, with $d=100$ and $d=128$ for CelebA and CIFAR-10, respectively. 
    We use 10,000 random training images and an equal number of random generated samples (decoded generated latent vectors) to compute FID. We use 200 random generated and 200 random real images for IS.
    } 
    \label{tab:comp_celeba_cifar_controlled}
\end{table*}







\begin{figure*}[h]
\centering
\includegraphics[width=1\textwidth]{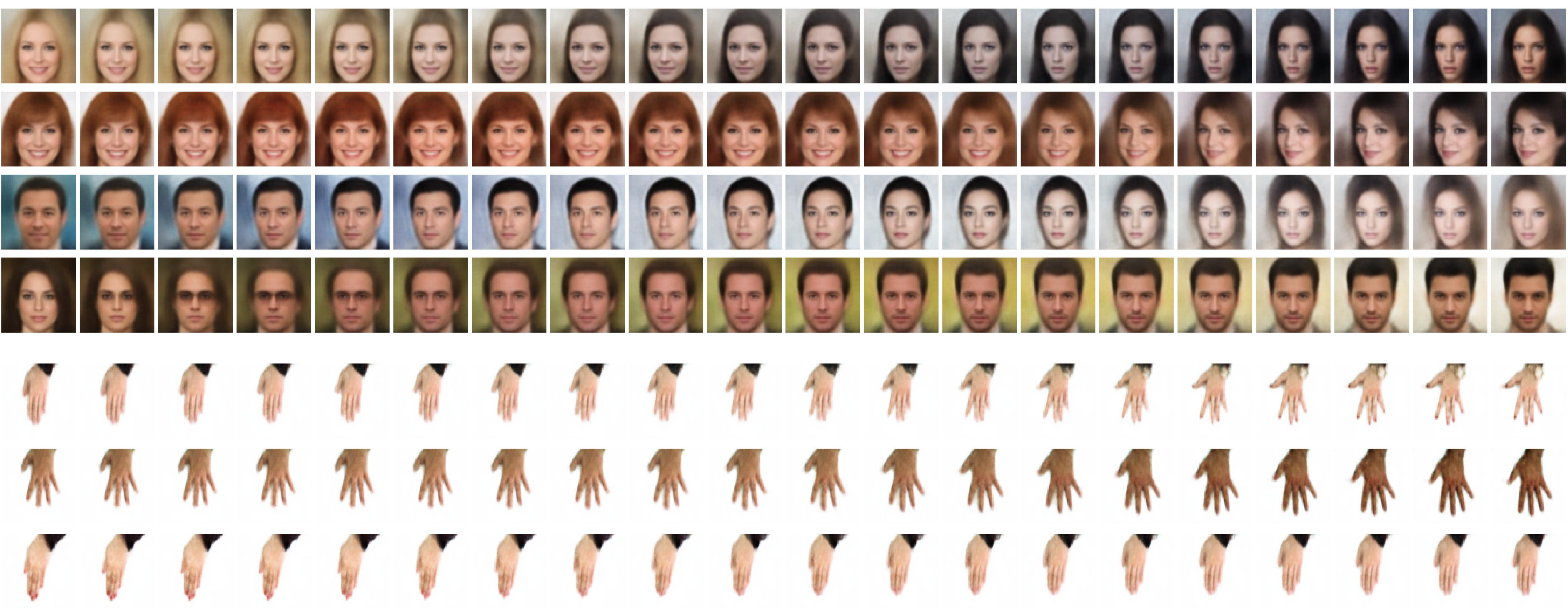}
\caption{Interpolations generated by a trained GTN for CelebA (top) and the HaP dataset (bottom).}
\label{fig:interp}
\end{figure*}

We next used the CelebA dataset as well as the Hands and Palm Images (HaP) dataset \cite{HaP} to demonstrate that GTNs indeed provide continuous interpolations. The HaP dataset is a small dataset containing $11,000$ images of upward-facing and downward-facing hands. We chose the HaP dataset since generating hands is known to be notoriously difficult for generative models \cite{handsNYT, handsYT}, let alone interpolating between them. As in CelebA, we trained a GTN for  each of the four dimensions, on images resized to $64\times 64$.

Figure \ref{fig:interp} shows that GTN indeed generates continuous interpolations. Each row contains generated interpolations between two  images (leftmost and rightmost). Specifically, each row contains the results of decoding $\hat{h}\big(\lambda y_{\text{left}} + (1-\lambda) y_{\text{right}}\big)$ for $20$ linearly spaced $\lambda \in [0,1]$, where, in CelebA, both $y_{\text{left}}$ and $y_{\text{right}}$ were sampled from $\mathcal{N}(\textbf{0},\textbf{I})$, and in HaP  $y_{\text{left}}$ and $y_{\text{right}}$ are each the normal label of a real encoded image (from Algorithm \ref{algorithm_labeling}), chosen so that they have the same orientation. 
For CelebA we used the $d=100$ model and for HaP we  used $d=50$ (chosen after consulting the IS and FID results across all four $d$ values in Appendix Table \ref{tab:IS_FID} while preferring a higher dimension ($d \in \{50, 100\}$) to retain more visual detail).





It is worth noting an interesting observation arising from the HaP dataset. Notice that the generated interpolations in Figure \ref{fig:interp} are between hands of the same orientation (downward facing). However, generated interpolations between hands (downward facing) and palms (upward-facing) do not seem as natural (Appendix Figure \ref{fig:interpolations_HaP_up_down}). This is likely because the dataset does not contain in-between positions for the latter (transitions between downward and upward), but it \textit{does} for the former (closed/open fingers to various degrees). This demonstrates that, regardless of methodology, the completeness of the dataset is important for accurate image generation, and particularly for interpolation. 


\section{Related work} \label{sec:related_work}

GTNs are related in concept to NFs
which seek to map a source distribution into a target distribution using a sequence of bijective transformations. These are typically implemented by a neural network, often in the form of at least tens of neural network blocks  \cite{xu2024embracing} and sometimes more \cite{xu2023mixflows}. 
The core of NFs is based on each of these transformations being a diffeomorphism
-- a specific type of homeomorphism that is more constrained than $h$ as it is defined between \textit{smooth} manifolds 
(as opposed to any topological spaces, including any manifolds) 
and requires that the function and its inverse are \textit{differentiable} (as opposed to just continuous for general homeomorphisms). NFs also require that the log-determinant of the Jacobian of these transformations is tractable, posing a constraint on the model architecture. 
GTNs do not pose any limitations on the model architecture. 
NFs also differ from GTNs in their optimization method since NFs employ a maximum likelihood objective -- typically the Kullback-Leibler (KL) divergence
while GTNs are trained using MSE. 
It was observed that optimizing the KL-divergence may be difficult for more complex normalizing flow structures \cite{xu2023mixflows}. NFs may also suffer from low-quality sample
generation \cite{behrmann2021understanding}, partially because the constraint on the Jacobian, besides limiting the model structure, may also lead to issues such as spurious local minima and numerical instability \cite{golinski2019improving}.

Continuous normalizing flow (CNF) are a relatively recent type of NFs that were developed to avoid the main constraints posed by NFs -- namely the requirements of invertibility and  a tractable Jacobian determinant. To avoid these constraints, CNFs use ordinary differential equations (ODEs) to describe the transformations in NFs. While CNFs have provided certain improvements over NFs, they are still time-intensive 
\cite{huang2021accelerating}. Improving both the speed and performance of CNFs is an active and promising field of research. 

Also closely related are VAEs since they are based on autoencoders that aim to transform the lower-dimensional latent space into a tractable distribution. The VAE loss function optimizes two terms, namely both the reconstruction error and the error between the prior and posterior distributions. This often leads to training instabilities and to posterior collapse. Despite these training instabilities, methods based on VAEs have been suggested, including Two-Stage VAE \cite{dai2019diagnosing} which first trains a VAE to identify the lower-dimensional manifold, and then a second VAE to transform the learned latent space into a normal distribution. Other methods,  combining VAE and flow, have also been suggested, but have seen limited success at improving upon VAE \cite{xiao2019generative} and add architectural limitations as mentioned earlier. Compared to VAE-based methods, GTNs have several practical advantages. One advantage is that GTNs separate between the autoencoder and the generative process, avoiding the intricacies of balancing between the two terms present in the VAE loss function (an in-depth analysis of the issues arising from this balancing can be found in \cite{dai2019diagnosing}). Practically, this means that one can focus on finding a high-quality latent representation of the data first, and then learn to generate in that latent space. 
This may be of particular interest in light of the use of pretrained autoencoders in state-of-the-art generative models \cite{rombach2022high}. 
Another advantage is the training stability, with no risk of posterior collapse, which  is a main issues with VAEs. 
Also worth noting is the fact that GTNs show no evidence of mode collapse (a known issue with other methods like GANs). 
These advantages, and the fact that GTNs employ a standard supervised learning approach, makes them more user-friendly and easier to train than many of the existing methods. 


\section{Discussion} \label{sec:discussion}

This work provides two main contributions: (1) It introduces a new class of generative models -- GTNs -- that offers training stability and simplicity; (2) It provides a new perspective on the importance of the intrinsic dimension of the data in the context of generative models.

As a generative method, GTNs offer a simple and stable approach to image generation. Compared to other generative methods like diffusion, VAE, GANs and NFs, GTNs are computationally simpler and more user-friendly since they employ a basic supervised learning approach that does not suffer from many of the training instabilities and specialised architectural requirements posed by other methods. GTNs also allow the flexibility of easily replacing the autoencoder should a better one be designed. This is of value in light of the recent turn to using pre-trained autoencoders for data generation \cite{rombach2022high}. VAEs, for example, would require retraining from scratch, which means risking dealing again with training instabilities and potential posterior collapse.

GTN's potential was demonstrated both quantitatively and qualitatively, particularly with demonstrable improvements over VAE. It is possible that with careful fine-tuning and with more sophisticated architectures to replace the fully-connected GTNs (e.g. to convolutional-based GTNs), further improvements may be attained (especially for CIFAR-10 where no attempts were made to fine-tune the settings or adapt the architecture when transitioning from CelebA). Upgrading from the vanilla autoencoders may also offer improvements, perhaps even more significant ones -- observing both the qualitative and quantitative similarity between reconstructions and generated samples (e.g. recon-IS being nearly achieved, the similarity between FID and real-FID as well as the similarities observed in Figure \ref{fig:cifar_celeb_mnist}) shows that a large portion of the error may be attributed to the quality of the autoencoder.

\begin{figure*}[ht!]
\centering
\includegraphics[width=1\textwidth]{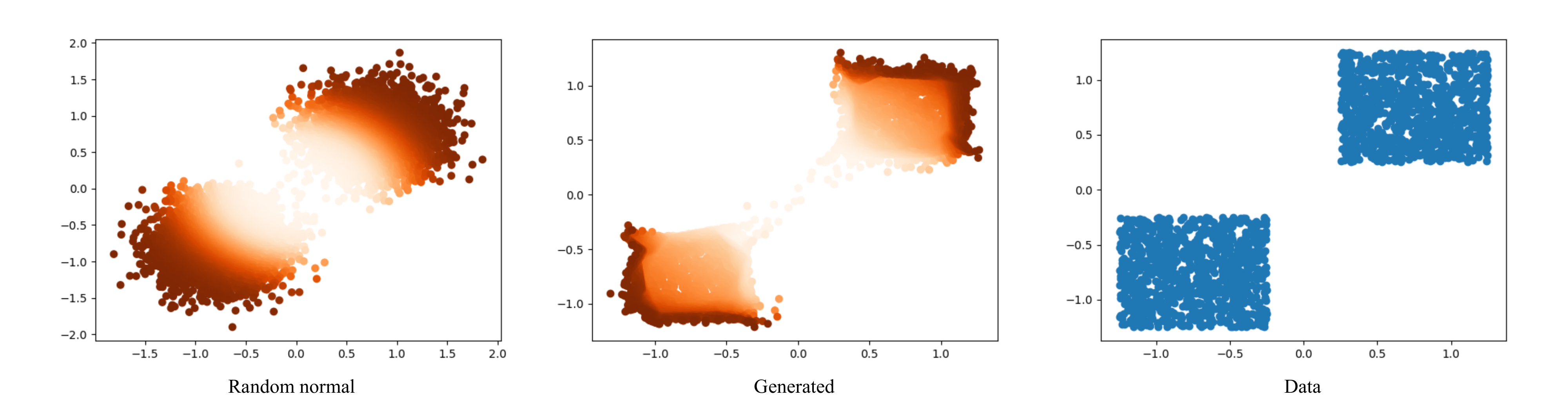}
\caption{Generating two disjoint uniform distributions. Left-to-right: mixture of random normal samples, corresponding predictions generated by a GTN and actual samples from the data. The random normal samples were generated by clustering the training data into two clusters using the fast-pytorch-kmeans package \cite{fastKmeans} and using the cluster means and standard-deviations to define the normal distributions. Labels were computed in each cluster separately according to Algorithm \ref{algorithm_labeling} to obtain a single dataset for training the GTN.}
\label{fig:disjoint_uniforms}
\end{figure*}

GTNs can also be extended to data that does not immediately satisfy the assumptions on the support of the 
distribution (like data that is separable into disconnected components).
In such a case, it is possible to exploit the assumption that the data lies on a manifold by labeling points 'locally' using a mixture model before training the GTN. We demonstrate this in Figure \ref{fig:disjoint_uniforms} where the data distribution is supported on two disjoint sets. Note that GTNs were also successfully applied to distributions that are not rotation invariant, particularly the uniform distribution. We leave a more thorough discussion and analysis of such extensions to future work. 

Finally, the topological considerations used throughout this paper, including those relating to the intrinsic dimension of the data, may prove useful for other methods, as already briefly discussed for diffusion. For example, NFs are by-definition sensitive to the same topological considerations discussed herein since they employ diffeomorphisms -- a strict form of homeomorphisms. In particular, this means that topological consideration like connectedness (Appendix \ref{appx_propertyTopo}) and the intrinsic dimension are important to consider there too. We note that particularly the intrinsic dimension of the data is an important property that is largely overlooked -- many generative methods have been used in the higher-dimensional pixel-space. Recently, latent generative methods have been developed and were shown to improve upon higher-dimensional ones \cite{rombach2022high}. One potential contribution to that success could lie in the fact that the latent dimension is closer to the intrinsic dimension of the data, reducing the risk of generating many outliers like those produced by diffusion in the provided swiss-roll example. More broadly, topological considerations like those used throughout this paper may prove useful for the design of better generative models and for the understanding of why some methods improve upon others.

\newpage
\clearpage 

\bibliography{aaai25}

\section*{Appendix}
\appendix

\section{}\label{appx_thm}
\begin{enumerate}
    \item $h$ is well-defined: it suffices to show that: (a) $F_X|_{S_X}$ is invertible, and (b) that the image of $F_Y|_{S_Y}$ is in the domain of $F_X|_{S_X}^{-1}$.
For (a), observe that $F_X|_{S_X}$ is continuous (the CDF of a continuous random variable is continuous)
, strictly monotonically increasing (the CDF is strictly monotonically increasing on the support) and onto $(0,1)$ (by definition of the CDF and the fact that the support is an open interval). As such it is invertible.
 A symmetric argument can be made for $F_Y|_{S_Y}$, which we will use later to define the inverse of $h$.

For (b), we know from the proof of (a) that the image of $F_Y|_{S_Y}$ is the domain of $F_X|_{S_X}^{-1}$.
\item $h$ is a homeomorphism since: a. it is continuous as a composition of continuous functions.
Indeed, $F_Y|_{S_Y}$ is continuous (the CDF of a continuous random variable is continuous). $F_X|_{S_X}^{-1}$ is also continuous: since $F_X|_{S_X}$ is continuous (as was $F_Y|_{S_Y}$) and bijective (it is invertible, as shown in the first part) we can use the known result that a continuous and bijective function between two open intervals has a continuous inverse (a consequence of the invariance of domain theorem). 
; b. it is bijective since the inverse is $F_Y|_{S_Y}^{-1}\circ F_X|_{S_X}$; c. its inverse is continuous: we know from a. and b. that $h$ is continuous and bijective so we can again use the fact that a continuous and bijective function between two open intervals has a continuous inverse.
\end{enumerate}


\section{}\label{appx_rationaleAlgo}

To understand the rationale behind Algorithm \ref{algorithm_labeling}, imagine four points from each of $X$ and $Y$ lying on the same line, say the $x$-axis in $\mathbb{R}^2$, such that both $X$ and $Y$ have two points on each side of the origin (we assume $D_X, D_Y$ are centralized). See Appendix Figure \ref{fig:illustration_algorithm_four_points_appx}, where circles are from $Y$ and stars are from $X$.
The cosine similarity of each pair of points $\{a, b\}$ with  $a\in D_X, b\in D_Y$ is $1$ if they are on the same side or $-1$ of they are on  opposite sides. Take the first $y \in D_Y^{\text{sorted}}$ (the closest one to the origin which reflects $\sim 50$th percentile since we assumed an equal number of points from $Y$ on both sides). The maximum cosine similarity out of all $x \in D_X^{\text{sorted}}$ is $1$, meaning $y$ will be matched with an $x$ on the same side. The fact that $D_X$ is sorted, means that we are using the distance as a tie breaker --
out of all $x \in D_X$ that are on the same side as $y$, the $x$ that is closest to the origin ($\sim 50$th percentile) will be chosen as $x_y$. Likewise, the leftmost $y$ will be matched with the leftmost $x$ (both $0$th percentile), the rightmost $y$ with the rightmost $x$ (both $100$th percentile) etc. Note that if this were in $\mathbb{R}$, this process coincides with the same labeling process described in the $1$D case (illustrated in Figure \ref{fig:labeling}). Note that if $D_X, D_Y$ are not sorted, discontinuities in the labeling may arise (see Figure \ref{fig:why_sorting_is_essential}).

\begin{figure}[ht!]
\centering
\includegraphics[width=1\columnwidth]{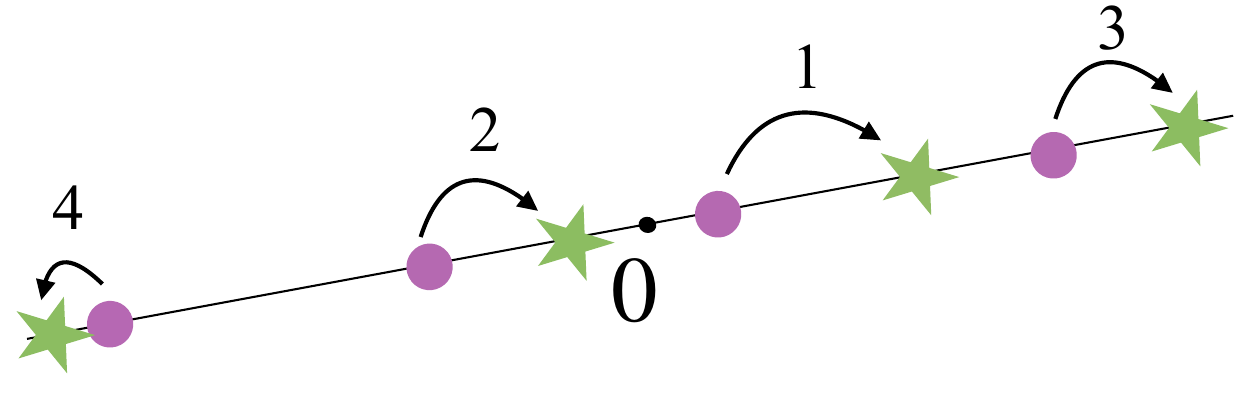}
\caption{Illustration of Algorithm \ref{algorithm_labeling} as described in Appendix \ref{appx_rationaleAlgo}. The numbers reflect the order in which the matching of $y$ with $x_y$ occurs.}
\label{fig:illustration_algorithm_four_points_appx}
\end{figure}

\begin{figure}[ht!]
\centering
\includegraphics[width=\columnwidth]{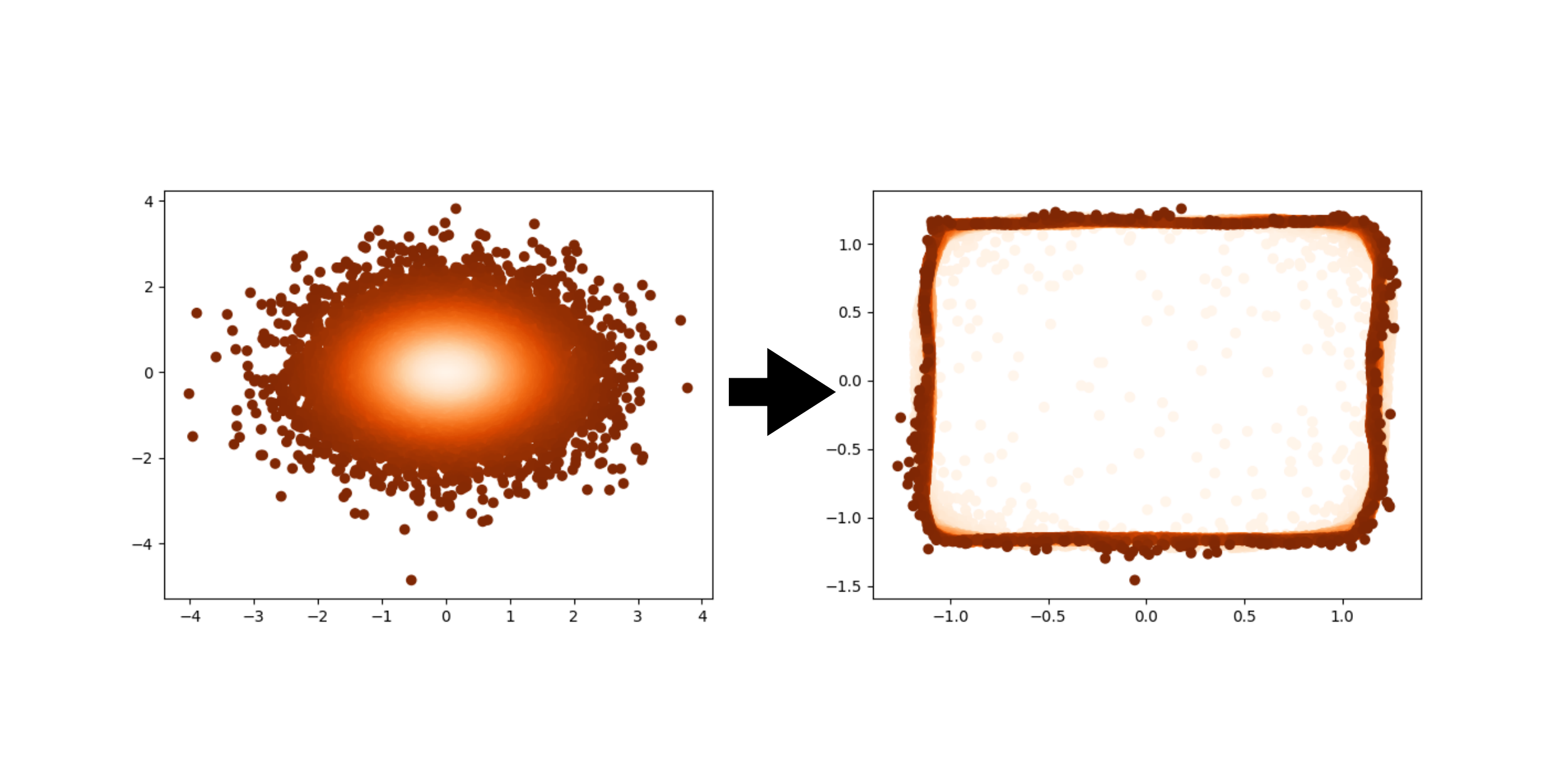}
\caption{Example demonstrating that sorting $D_X, D_Y$ in the labeling algorithm is essential. Training GTN on the 2D-uniform data on labels produced by the same algorithm with sorting removed causes test predictions to collapse mainly to the boundary.}
\label{fig:why_sorting_is_essential}
\end{figure}

\section{}\label{appx_propertyTopo}
One example of a topological property is the connectedness property -- being a "single piece", like $\mathbb{R}$, as opposed to "more than one piece", like $(-\infty,-1)\cup (1, \infty)$, is preserved between homeomorphic spaces (so, in particular, these two are not homeomorphic, but $\mathbb{R}$ \textit{is} homeomorphic to each of the two intervals, separately). Therefore, if we suspect that our data is composed of separate classes like images of both hands and faces, and we would like to train a GTN as a generative model, one option would be to use a clustering approach and to generate labels for each cluster separately. Another would be to split our data into its different components (learning a separate $h$ for each). This relates to the conditions of Theorem \ref{thm} which assume that $X$'s support is an open interval. For example, if we suspect that our data can be separated into two disjoint uniform distributions, then the theorem doesn't apply. However, it \textit{does} apply to each separate uniformly distributed component.  




\begin{figure*}[ht!]
\centering
\includegraphics[width=1\textwidth]{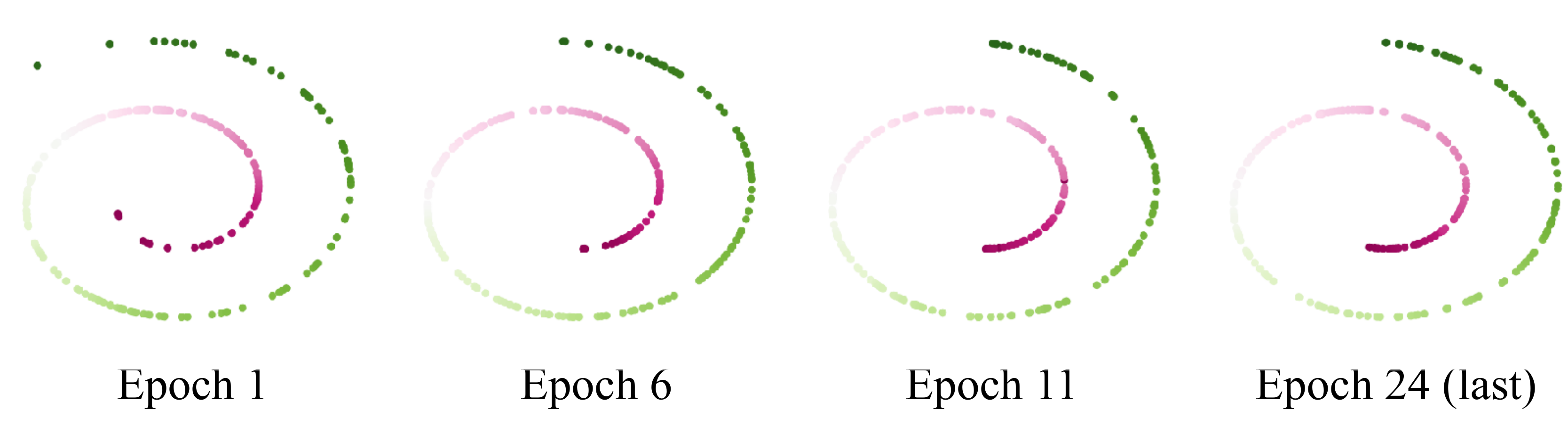}
\caption{Swiss-roll samples generated by GTN during the training process.}
\label{fig:swiss_epochs}
\end{figure*}

\begin{figure*}[ht!]
\centering
\includegraphics[width=1\textwidth]{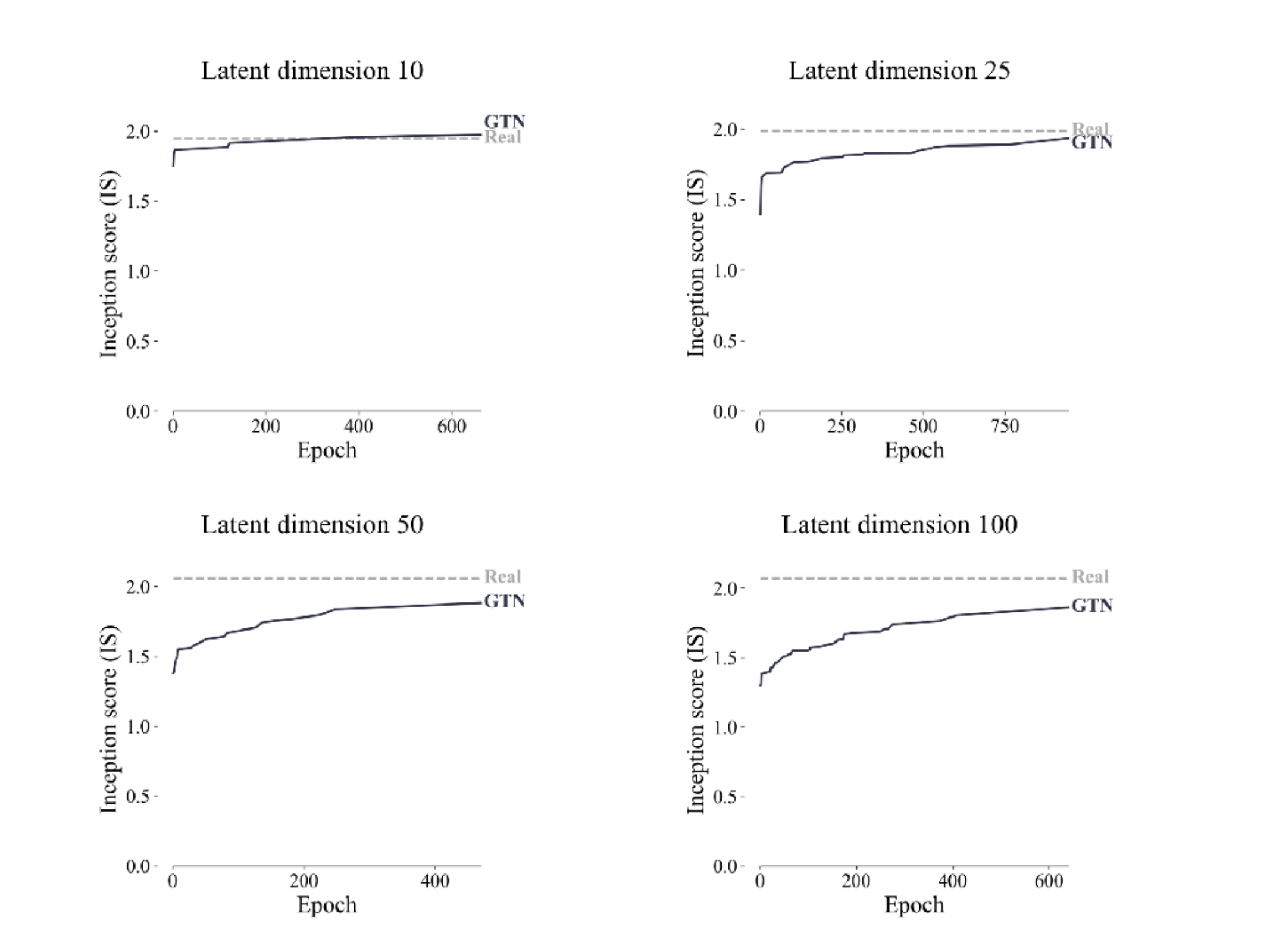}
\caption{Inception score (IS) vs recon-IS for CelebA plotted by epoch (at epochs of improvement in the IS). At each epoch, the IS was computed for $200$ randomly generated images (decoded latent vectors generated by $\hat{h}$ from random normal vectors, formally: $autoencoder.decoder\big(\hat{h}(r)\big)$ for $r \sim \mathcal{N}(\textbf{0},\textbf{I})$)  (solid black line). recon-IS  was computed once for a random set of $200$ reconstructions (dashed gray line).}
\label{fig:epoch_to_IS_grid_CelebA}
\end{figure*}

\begin{figure*}[ht!]
\centering
\includegraphics[width=1\textwidth]{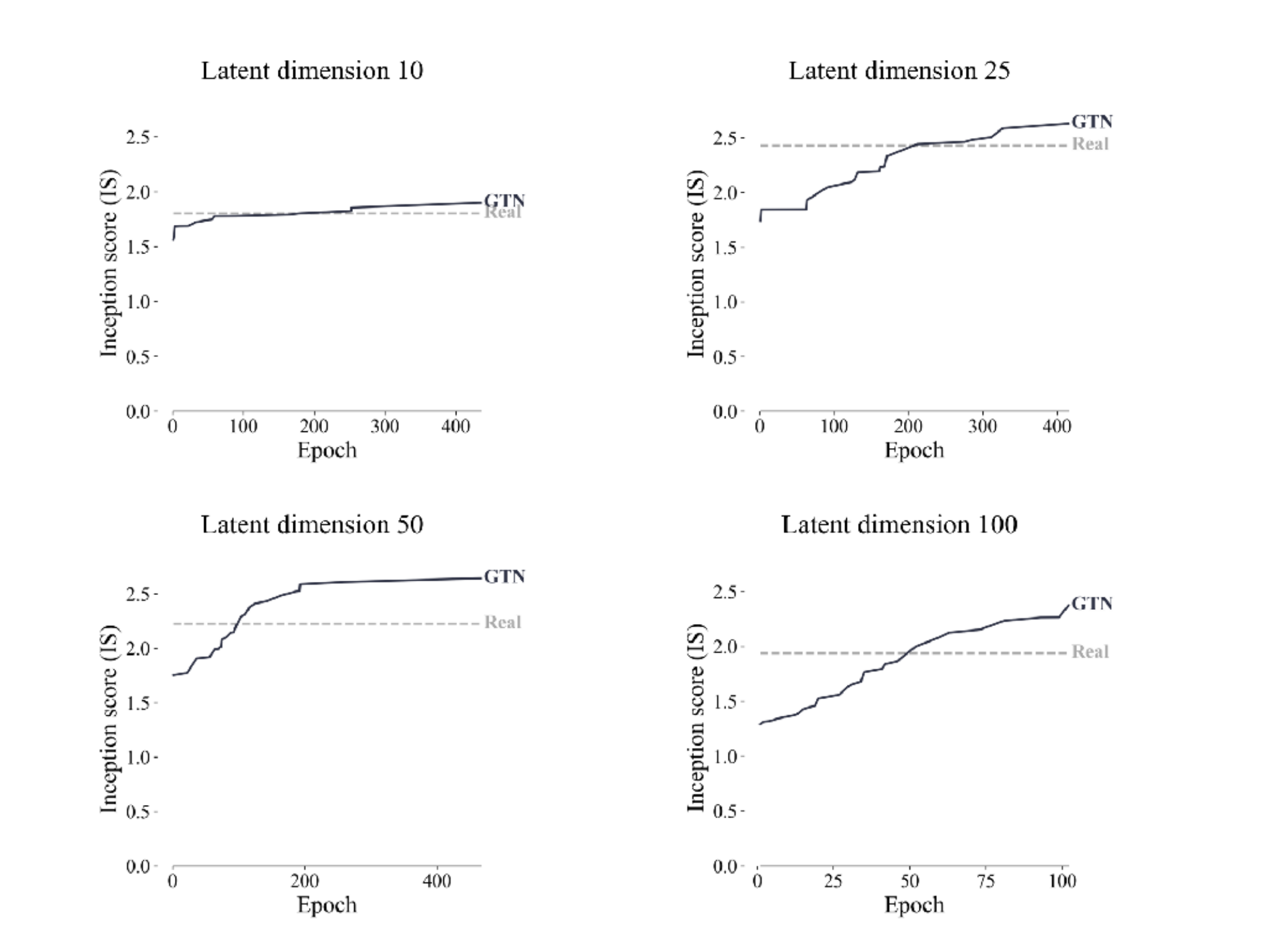}
\caption{Inception score (IS) vs recon-IS for HaP plotted by epoch (at epochs of improvement in IS). At each epoch, IS was computed for $200$ randomly generated images (decoded latent vectors generated by $\hat{h}$ from random normal vectors), formally: $autoencoder.decoder\big(\hat{h}(r)\big)$ for $r \sim \mathcal{N}(\textbf{0},\textbf{I})$) (solid black line). recon-IS  was computed once for a random set of $200$ reconstructions (dashed gray line).}
\label{fig:epoch_to_IS_grid_HaP}
\end{figure*}

\begin{figure*}[ht!]
\centering
\includegraphics[width=1\textwidth]{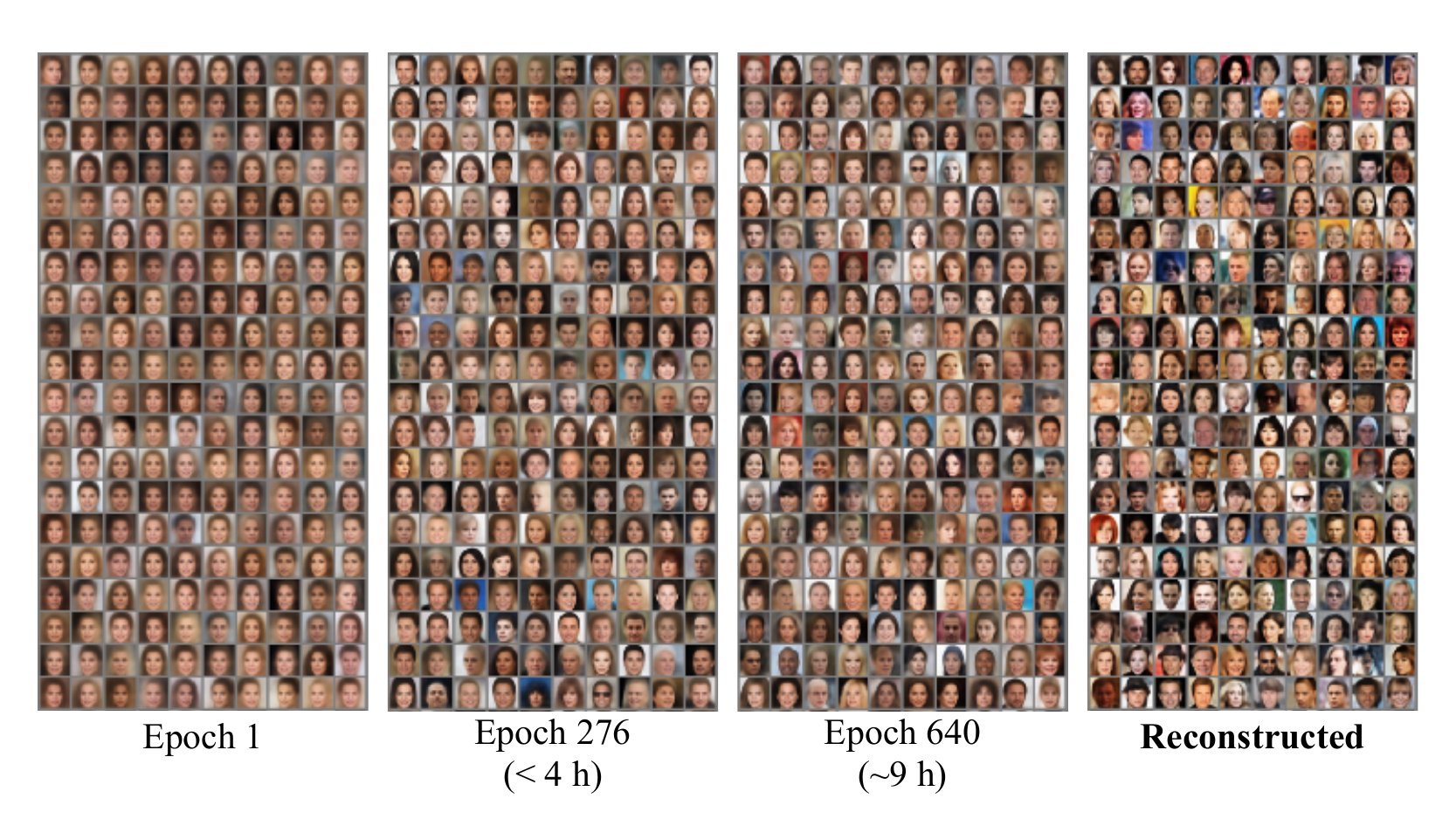}
\caption{Randomly chosen samples generated by a GTN during the training process on CelebA with latent dimension 100 at several epochs of improvement in the IS score, and randomly chosen real images (bottom-right grid). Specifically, each epoch shows 200 images, each of which is the decoded $\hat{h}(r)$ for some random $r \sim \mathcal{N}(\textbf{0},\textbf{I})$). The bottom right grid shows 200 images each of which is the decoded latent vector of a random real image. Training until epoch 640 took approximately $9$ hours on a single T4 GPU, with similar-quality images obtained in less than $4$ hours of training at epoch $276$.}
\label{fig:celeba_epochs}
\end{figure*}

\begin{figure*}[ht!]
\centering
\includegraphics[width=\textwidth]{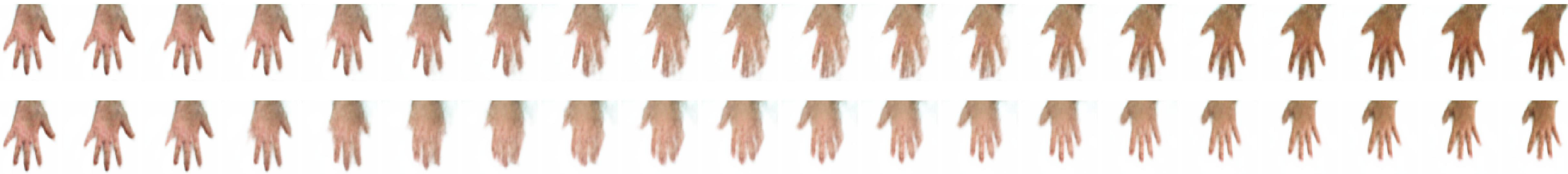}
\caption{Interpolations between a hand (downward facing) and a palm (upward facing).}
\label{fig:interpolations_HaP_up_down}
\end{figure*}

\section{}\label{appx_thm2}

    $h$ is injective:  equating for two  points $y_1, y_2$ we obtain: $$ h_1(||y_1||)\frac{y_1}{||y_1||} = h_1(||y_2||)\frac{y_2}{||y_2||} $$ Applying the norm to both sides yields $h_1(||y_1||) = h_1(||y_2||)$.  Since $h_1$ is injective, this means that $||y_1||=||y_2||$. Plugging this into the equality and cancelling out equal terms ($y_1,y_2 \neq 0)$ yields $y_1=y_2$. The function is also onto: since $h_1$ is onto (it is a homeomorphism) then for any $x\neq 0$ there is a value $v$ in $(0,\infty)$ with $h_1(v)=||x||$. Since the support of $Y$ is $\mathbb{R}^d$ there is a $y$ that is on the same line as $x$ from the origin and that satisfies $||y||=v$ so that: $x=h_1(||y||)\frac{y}{||y||}$, meaning that $x$ has a $y \in S_Y$ for which $h(y)=x$. 
    
    $h$ is continuous since for $y\neq 0$ it is a composition of continuous functions, and for $y=0$ we observe that $\lim_{y\to 0}  h_1(||y||)\frac{y}{||y||} = 0$ ($\frac{y}{||y||}$ is the unit vector in the direction of $y$ and $h_1(||y||) \to 0$ as $y \to 0$ by definition of the CDF).

    The inverse is also continuous since a continuous and bijective functions between open subsets of $\mathbb{R}^d$ is continuous (by the invariance of domain theorem).

\section{}\label{appx_another_way_lines}
    Another way of thinking about this is as follows: assume for simplicity that $f_X, f_Y > 0$ (so that $S_X, S_Y = \mathbb{R}^d$ and therefore  $l\cap S_X$ and $l\cap S_Y$ are simply $l$).
Since $l$ is a $1$D manifold, $X$ and $Y$ induce $1$D random variables $X^l$ and $Y^l$ on the line $l$, with 
pdfs $f_{X^l}, f_{Y^l}$ and CDFs $F_{X^l}, F_{Y^l}$. We can  apply Theorem \ref{thm} to these random variables to obtain a homeomorphism $h^l$ for each line. 

\section{}\label{appx_architectures_and_training}
The same autoencoder architecture was used for MNIST, CelebA and HaP with only the essential modifications needed to accomodate to the different input channels (1 for MNIST and 3 for CelebA and HaP) and the different latent dimensions. The architecture consisted of a decoder with two 2d convolutional layers followed by ReLU activation: the first convolutional layer had 64 output channels, kernel size 4, stride 2 and padding 1. The second convolutional layer consisted of 128 output channels, kernel size 4, stride 2 and padding 1. The two convolutional layers were followed by a liner layer with the number of output features set to the desired latent dimension $d$. The output activation was tanh. In the decoder we used a mirror architecutre of one linear layer, with the number of input features being $d$, and two 2d transposed convolution layers. The code for all architectures is also available in our repository. To train the autoencoders we used: a batch size of $200$ for CelebA and HaP, and $128$ for MNIST; learning rate $1e-4$ for CelebA and HaP and $1e^{-3}$ for MNIST; a weight decay of $1e^{-5}$ was used in all three.
We tested the two learning rates $1e^{-3}$ and $1e^{-4}$. No further optimizations were made for the autoencoder hyperparameters.

CelebA images were center-cropped to $148 \times 148$ and resized to $64 \times 64$. HaP images were resized to $64 \times 64$. 

We did not use any data augmentations during the training process.

Final architecture specifications for $\hat{h}$ appear in Appendix Table \ref{tab:architectures_h_hat}. We ran a hyperparameter search for CelebA and HaP. Initially, we ran several settings for CelebA using a small number of epochs (roughly 100-300) for latent dimension $200$ prior to consulting the literature on the intrinsic dimension of image data. We tried learning rates of $1e^{-3}, 1e^{-4}, 1e^{-5}, 5e^{-5}$ for various width and depth settings. Specifically, for CelebA we tested widths of $500, 1000, 1200, 1300, 1500$ and depths of $10, 17, 20, 25, 27$. For HaP we kept the depth at $25$ after observing this obtained best results for CelebA and tested widths of $2000, 3000$ as well after identifying that HaP benefited from higher width settings. 

The architectures were compared based on their IS score at each epoch. The architecture with the highest IS score was kept. These architectures are the ones described in Appendix Table \ref{tab:architectures_h_hat}.

We used the best settings, shown in Appendix Table \ref{tab:architectures_h_hat}, regardless of the dimension $d$. 
We did not perform architecture optimizations per dimension. Such optimizations may provide further improvements in efficiency and/or generative quality.

Each experiment was run on 1 NVIDIA T4 GPU with 8 vCPU + 52 GB memory, 500GB SSD. Specifically, we used the "Deep Learning VM" by Google Click to Deploy in the Google Cloud Marketplace (image: pytorch-1-13-cu113-v20230925-debian-10-py37), modified to the aforementioned specifications.

\section{}\label{appx_swiss_1d_method}
The 1D swiss-roll data is generated by sampling $\theta \sim U(1.5\pi, 4.5\pi)$ and computing $f(\theta)=\theta(\cos{\theta}, \sin{\theta})$. GTN is trained to generate $\theta$. This example, besides demonstrating the 1D case, serves to demonstrate how GTN operates on the 'correct' latent representation of the data.

\section{}\label{sec:significance}
The fact that $h$ is a homeomorphism is significant in the context of generative models for several reasons: 

1. \textbf{Learnability} -- Since $h$ being a homeomorphism implies that both $h$ and $h^{-1}$ are continuous real-valued functions over some subset of $\mathbb{R}$, then by the universal approximation theorem they can be approximated to arbitrary accuracy by a neural network \cite{hornik1989multilayer}. \textit{This is demonstrated in Figure \ref{fig:intro_methods} (A).}

2. \textbf{Coverage and diversity} -- The bijectivity of a homeomorphism means that there is a one-to-one and onto correspondence between the supports $S_X$ and $S_Y$. This means that we can use $Y$ to cover all samples that can be obtained from $X$ and that no two samples in $Y$ will generate the same sample in $X$. \textit{This is demonstrated in Figure \ref{fig:intro_methods} (A) and (B).}

3. \textbf{Continuous interpolation} -- The fact that $h$ is continuous is significant for purposes of interpolation. 
For example, given a generative model $g:\mathbb{R} \to S_X$, two points $y_1, y_2 \in \mathbb{R}$, and the function: $\phi(\lambda)=\lambda y_1 + (1-\lambda)y_2$ (where $\phi: [0,1] \to \mathbb{R}$) we would like $(g\circ \phi)(\lambda)$ to be 
continuous (e.g. for video generation). 
If $g$ is stochastic, for example, this cannot be guaranteed. However, using $h$ as $g$, we have that $g\circ \phi$ is continuous as a composition of continuous functions. 
This provides the desired continuous interpolation between points from $X$. \textit{This is demonstrated in Figure \ref{fig:interp}.}

4. \textbf{Guiding topological properties} -- There are useful properties that are invariant under homeomorphisms ("topological properties") that can guide us in designing better generative models. For example, one property is  that homeomorphic manifolds must have the same dimension. \textit{The use of this property is  demonstrated in the upcoming swiss-roll example}, where it will lead us to the conclusion that we are better off generating swiss-roll samples from a $1$D standard normal distribution, rather than from a $2$D one. 
Another example for a useful property is in Appendix \ref{appx_propertyTopo}.

\begin{table*}[]
\begin{tabular}{lllllllllll}
\toprule
           & \textbf{}                                                             & \textbf{}      & \textbf{}                                                 & \textbf{}                                                               & \textbf{}                                                         & \textbf{}                                                        & \textbf{}          &           &           \\
           & \textbf{\begin{tabular}[c]{@{}l@{}}No. Hidden \\ Layers\end{tabular}} & \textbf{Width} & \textbf{Activation}                                       & \textbf{\begin{tabular}[c]{@{}l@{}}Batch \\ Norm.\end{tabular}} & \textbf{\begin{tabular}[c]{@{}l@{}}Learning \\ Rate\end{tabular}} & \textbf{\begin{tabular}[c]{@{}l@{}}Weight \\ Decay\end{tabular}} & \textbf{Optimizer} & \textbf{\begin{tabular}[c]{@{}l@{}}Batch \\ Size\end{tabular}} &\textbf{} & \textbf{} \\ \cline{2-9}
           &                                                                       &                &                                                           &                                                                         &                                                                   &                                                                  &                    &           &           \\
Swiss-Roll & 4                                                                     & 6              & \begin{tabular}[c]{@{}l@{}}LeakyReLU(0.5)\end{tabular} & No                                                                      & $1e^{-3}$                                        & No                                                               & Adam    & 250           &           &           \\
Uniform    & 6                                                                     & 6              & \begin{tabular}[c]{@{}l@{}}LeakyReLU(0.5)\end{tabular} & No                                                                      & $1e^{-3}$                                        & No                                                               & Adam   & 250            &           &           \\
MNIST      & 7                                                                     & 50             & \begin{tabular}[c]{@{}l@{}}LeakyReLU(0.5)\end{tabular} & Yes                                                                     & $1e^{-3}$                                         & No                                                               & Adam    & 128           &           &           \\
CelebA     & 26                                                                    & 1,200          & \begin{tabular}[c]{@{}l@{}}LeakyReLU(0.5)\end{tabular} & Yes                                                                     & $5e^{-5}$                                        & No                                                               & Adam  & 200              &           &           \\
HaP        & 26                                                                    & 3,000          & \begin{tabular}[c]{@{}l@{}}LeakyReLU(0.5)\end{tabular} & Yes                                                                     & $5e^{-5}$                                        & No                                                               & Adam    & 200          &           &           \\
CIFAR-10        & 26                                                                    & 1,500          & \begin{tabular}[c]{@{}l@{}}LeakyReLU(0.5)\end{tabular} & Yes                                                                     & $5e^{-5}$                                        & No                                                               & Adam    & 200          &           &           \\
           &                                                                       &                &                                                           &                                                                         &                                                                   &                                                                  &                    &           &           \\ 
\bottomrule
\end{tabular}
\caption{Architecture specifications for $\hat{h}$ for the different datasets. The architecture was kept the same between the different dimensions where different dimensions were tested (CelebA, HaP). } \label{tab:architectures_h_hat}
\end{table*}


\begin{table*}\centering
\begin{tabular}{@{}lrrrrcrrrcrrr@{}}\toprule
& \multicolumn{3}{c}{\textbf{CelebA}} & \phantom{abc}& \multicolumn{3}{c}{\textbf{HaP}} &
\phantom{abc} \\
\cmidrule{2-4} \cmidrule{6-8}
\\
 & \textbf{IS} $\uparrow$ &\textbf{ recon-IS} $\uparrow$ & \textbf{FID} $\downarrow$  & \phantom{abc}& \textbf{IS} $\uparrow$ & \textbf{recon-IS} $\uparrow$ & \textbf{FID} $\downarrow$ & \phantom{abc}\\
\cmidrule{2-4} \cmidrule{6-8} \\
$d=10$ & 1.97 & 1.95  & 119.46   &&   1.90& 1.80 & 192.98 & \\
$d=25$ & 1.93 & 1.99 & 82.88  &&   2.63& 2.43& 123.89 &\\
$d=50$ & 1.88 & 2.06 & 71.61  &&   2.64& 2.22& 81.48\phantom{a}&\\
$d=100$ & 1.86 & 2.07 & 66.05   &&   2.37& 1.94& 101.85 &\\
\bottomrule
\end{tabular}
\caption{Results per dataset and latent dimension using vanilla autoencoders (2 layer encoder). Inception Score (IS) is the best IS attained during training for the given dimension and dataset for a random set of 200 reconstructions. recon-IS is the IS attained by a set of 200 reconstructions. Fréchet Inception Distance (FID) is computed using $50,000$ randomly generated images and $50,000$ random real images
using the same model that attained the reported best IS.  For the smaller HaP dataset, all training images  ($\sim$ 8,000) and the same number of randomly generated images were used.} \label{tab:IS_FID}
\end{table*}

\begin{table*}
    \centering
    \begin{tabular}{ cc }
    \toprule
    \textbf{Method} &  \textbf{FID $\downarrow$}  \\
     \midrule
     2-Stage VAE \cite{dai2019diagnosing} &  44.4  \\
     NVAE \cite{vahdat2020nvae} & 14.74 \\
     WAE-GAN \cite{tolstikhin2017wasserstein} & 42.0 \\
     DiffuseVAE \cite{pandey2022diffusevae} & 3.97 \\
     GLF \cite{xiao2019generative} & 53.2 \\
     VAE + flow posterior \cite{xiao2019generative} & 67.9 \\
     VAE + GMM=75 \cite{pandey2022diffusevae} & 72.11 \\
      GTN (ours) & 66.05  \\ 
     \bottomrule
    \end{tabular}
    \caption{GTN results and results reported by related methods on CelebA. GTN results are for $d=100$ (from Table \ref{tab:IS_FID}). 
    All other results are as originally reported by their authors, except for NVAE which does not report FID (results are as reported by \cite{pandey2022diffusevae}). Note that the methods may differ in their evaluation settings and implementation. 
    } \label{tab:comp_celeba_many}
\end{table*}

\begin{table*}
    \centering
    \centering
    \begin{tabular}{ cc }
    \toprule
    \textbf{Method} &  \textbf{FID $\downarrow$}  \\
     \midrule
     2-Stage VAE \cite{dai2019diagnosing} &  72.9  \\
     NVAE \cite{vahdat2020nvae} & 51.67 \\
     DiffuseVAE \cite{pandey2022diffusevae} & 2.62 \\
     GLF \cite{xiao2019generative} & 88.3 \\
     VAE + flow posterior \cite{xiao2019generative} & 143.6 \\
     VAE + GMM=75 \cite{pandey2022diffusevae} & 137.68 \\
      GTN (ours) & 238.62  \\ 
     \bottomrule
    \end{tabular}
    \caption{GTN results and results reported by related methods on CIFAR-10. GTN results are from Table \ref{tab:comp_celeba_cifar_controlled}. 
    All other results are as originally reported by their authors, except for NVAE which does not report FID (results are as reported by \cite{pandey2022diffusevae}). Note that the methods may differ in their evaluation settings and implementation.
    } \label{tab:comp_cifar_many}
\end{table*}

\begin{figure*}[ht!]
\centering
\includegraphics[width=1\textwidth]{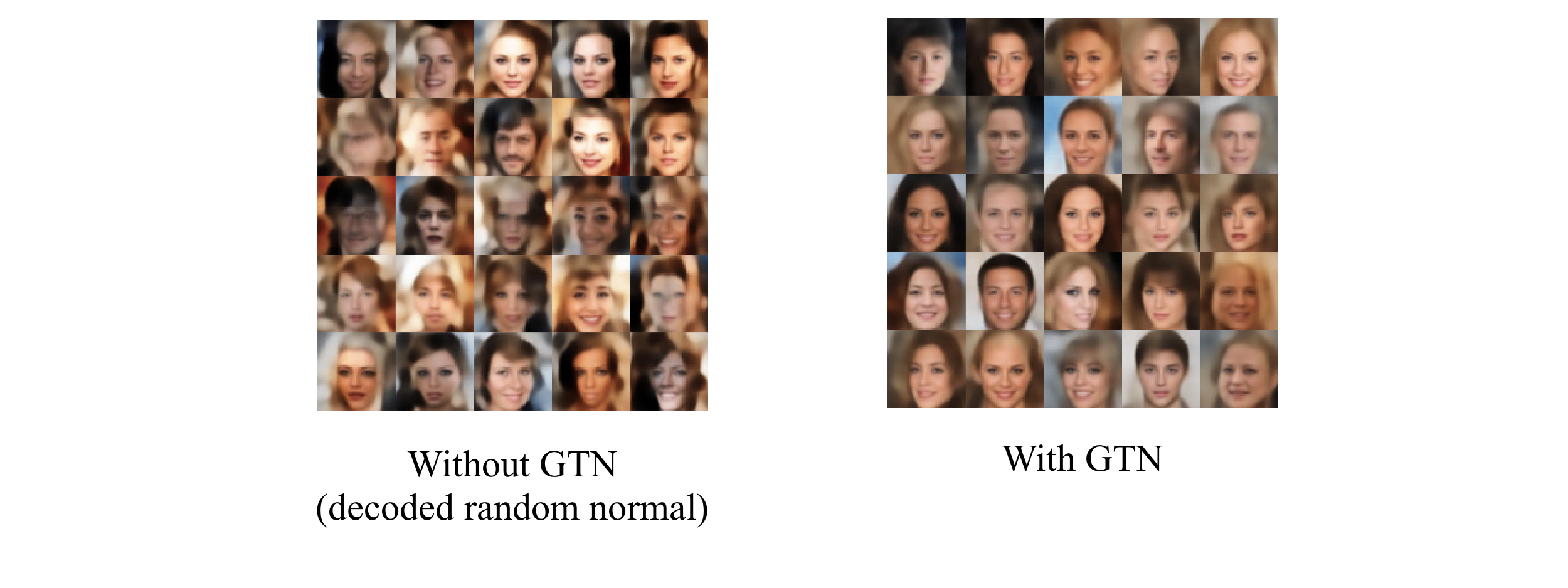}
\caption{Generated images by: (left) decoding random normal vectors with vectors $\mu, \sigma$ estimated from the latent space (each image is the decoded $r$ for some random $r \sim \mathcal{N}(\mu,\sigma I)$); (right) GTN (each image is the decoded $\hat{h}(r)$ for some random $r \sim \mathcal{N}(0,I)$)).}
\label{fig:random_normal_vs_gen}
\end{figure*}
\end{document}